\pdfoutput=1
\PassOptionsToPackage{backref=page}{hyperref}
\documentclass[11pt]{article}

\newif\ifauthordecided
\authordecidedtrue

\newif\ifarxiv
\arxivtrue %

\newif\ifonepair
\onepairfalse

\newif\ifperfect
\perfectfalse %

\ifarxiv
    \usepackage{acl}
\else 
    \usepackage[review]{acl}
\fi

\usepackage{times}
\usepackage{latexsym}

\usepackage[T1]{fontenc}

\usepackage[utf8]{inputenc}

\usepackage{microtype}

\usepackage{sec/package} %

\usepackage{amsmath}
\usepackage{units}

\newcommand{\tfact}{t_{\mathrm{fact}}\xspace}
\newcommand{\talt}{t_{\mathrm{cofa}}\xspace}
\newcommand{\dalt}{\Delta_{\mathrm{cofa}}\xspace}

\usepackage{tcolorbox}

\title{
Competition of Mechanisms: \\
Tracing How Language Models Handle Facts and Counterfactuals}

\ifauthordecided
\author{Francesco Ortu\thanks{\hspace{0.1cm} Equal contributions.}
\\
  University of Trieste \\
  \small\texttt{francesco.ortu@studenti.units.it
} \\\And
  Zhijing Jin\samethanks \\
  MPI \& ETH Zürich \\
  \small\texttt{jinzhi@ethz.ch} \\ \And
   Diego Doimo\\
   AREA Science Park \\
  \small\texttt{diego.doimo@areasciencepark.it} \\ \AND 
  Mrinmaya Sachan \\
  ETH Zürich \\
  \small\texttt{msachan@ethz.ch} \\ \And
  Alberto Cazzaniga\thanks{\hspace{0.1cm} Co-supervision.}\\
   AREA Science Park \\   \small\texttt{alberto.cazzaniga@areasciencepark.it} \\\And
  Bernhard Sch\"olkopf\samethanks \\
  MPI for Intelligent in Systems \\
  \small\texttt{bs@tue.mpg.de}\\
}
\fi

\begin{document}

\maketitle
\begin{abstract}
Interpretability research aims to bridge the gap between empirical success and our scientific understanding of the inner workings of large language models (LLMs).
However, most existing research focuses on analyzing a single mechanism, such as how models copy or recall factual knowledge. In this work, we propose a formulation of \textit{competition of mechanisms}, which focuses on the interplay of multiple mechanisms instead of individual mechanisms and traces how one of them becomes dominant in the final prediction.
We uncover how and where mechanisms compete within LLMs using two interpretability methods: logit inspection and attention modification. Our findings show traces of the mechanisms and their competition across various model components and reveal attention positions that effectively control the strength of certain mechanisms.%
\footnote{
\ifarxiv
Code: {\href{https://github.com/francescortu/comp-mech}{https://github.com/francescortu/comp-mech}. Data: \href{https://huggingface.co/datasets/francescortu/comp-mech}{https://huggingface.co/datasets/francescortu/comp-mech}}.
\fi
}
\end{abstract}

\section{Introduction}
Recent advancements in large language models (LLMs) have brought unprecedented performance improvements to NLP \cite[\textit{inter alia}]{brown2020gpt3,touvron2023llama,openai2023gpt4,anil2023gemini}. However, the black-box nature of these models obfuscates our scientific understanding of \textit{how these models achieve certain capabilities}, and \textit{how can we trace the problem when they fail at other tasks}.
This has brought an increasing focus on interpretability research to help us understand the inner workings of LLMs.
\begin{figure}[t]
    \centering
        \centering
        \includegraphics[width=\linewidth]{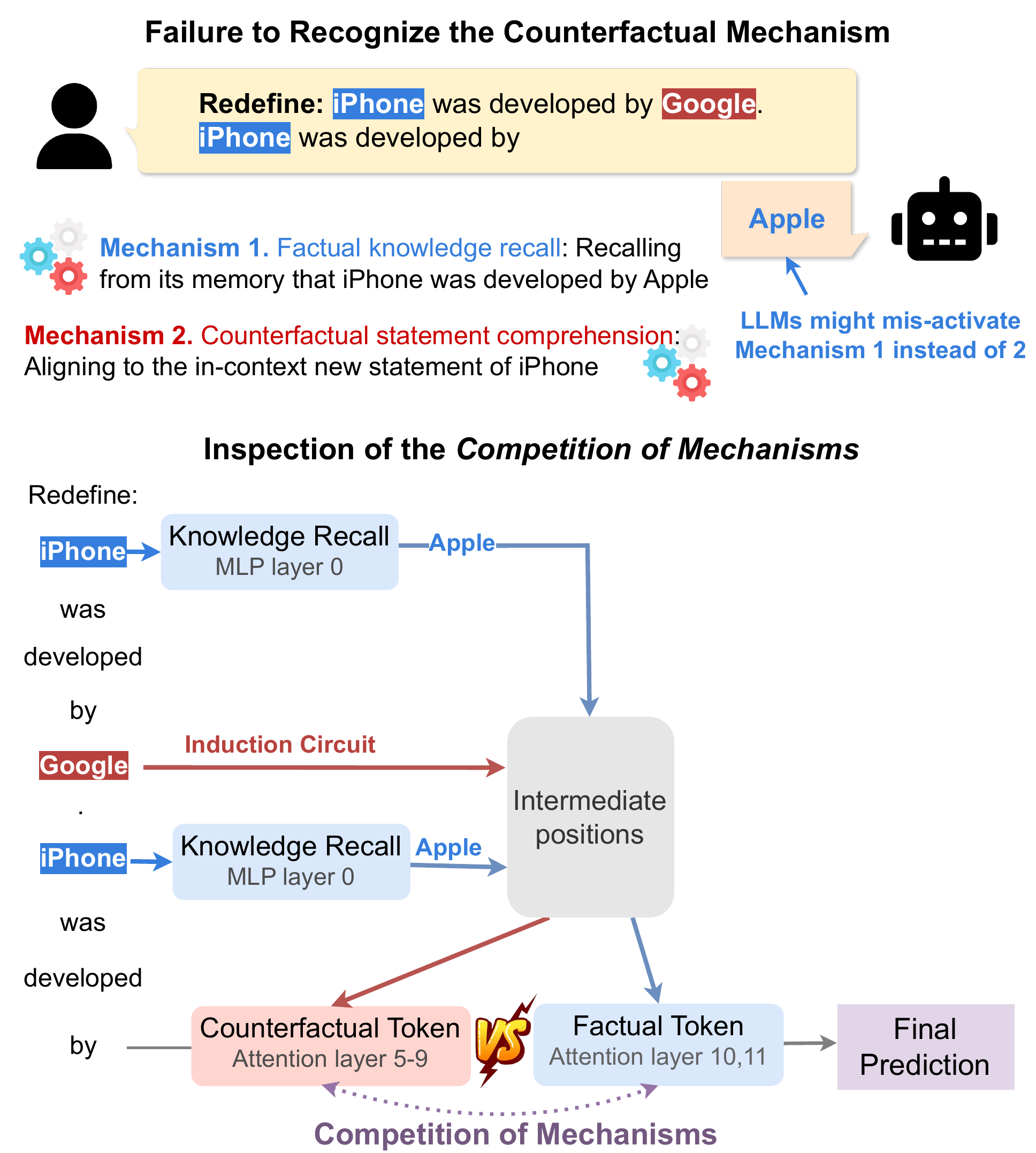}
    \caption{Top: An example showing that LLMs can fail to recognize the correct mechanism when multiple possible mechanisms exist.
    Bottom: Our mechanistic inspection of where and how the competition of mechanisms takes place within the LLMs.
    }\label{fig:examples}
\end{figure}

Existing interpretability research has been largely focused on discovering the \textit{existence} of single mechanisms, such as the copy mechanism in induction heads of LLMs \cite{elhage2021mathematical,olsson2022incontext}, and factual knowledge recall in the MLP layers \citep{geva-etal-2021-transformer,meng2022locating,geva2023dissecting}. However, different from discovering \textit{what mechanisms exist in LLMs}, we propose a more fundamental question: \textit{how do different mechanisms interact in the decision-making of LLMs?}
We show a motivating example in \cref{fig:examples},
where the model fails to recognize the correct mechanism when it needs to judge between two possible mechanisms: whether to recall the factual knowledge on who developed the iPhone (i.e., Mechanism 1) or to follow its counterfactual redefinition in the new given context (i.e., Mechanism 2).

We propose a novel formulation of \textit{competition of mechanisms}, which focuses on tracing each mechanism in the model and understanding how one of them becomes dominant in the final prediction by winning the ``competition''. Specifically, we build our work on two single mechanisms that are well-studied separately in literature: (1) the factual knowledge recall mechanism, which can be located in the MLP layers \citep{geva-etal-2021-transformer,meng2022locating,geva2023dissecting}; and (2) the in-context adaptation to a counterfactual statement, which is enabled by the copy mechanism conducted by induction heads of attention layers \cite{elhage2021mathematical,olsson2022incontext}.
Based on the latest tools to inspect each of these two mechanisms \cite{logitlens2020lesswrong,wang2023IOI,geva2023dissecting}, we then unfold \textit{how and where} the competition of the two mechanisms happen, and how it leads to the overall success or failure of LLMs.

Technically, we deploy two main methods: logit inspection \cite{logitlens2020lesswrong,geva-etal-2022-transformer} by projecting the outputs of each model component by an unembedding matrix, and attention modification 
\cite{geva2023dissecting,wang2023IOI}.
Using these methods, we assess the contributions of various model components, both from a macroscopic view (e.g., each layer) and a microscopic view (e.g., attention heads), and identify
critical positions and attention heads involved in the competition of the two mechanisms.
Moreover, we locate a few localized positions of some attention head matrices that can significantly control the strength of the factual mechanism. 
We summarize our main findings as follows:

\begin{enumerate}
    \item In early layers, the factual attribute is encoded in the subject position, while the counterfactual is in the attribute position (\cref{sec:residual_stream});
    \item The attention blocks write most of the factual and counterfactual information to the last position (\cref{sec:attn_mlp_contrbution});
    \item All the highly activated heads attend to the attribute position regardless of the specific type of information they promote. The factual information flows by penalizing the counterfactual attribute rather than promoting the factual one (\cref{sec:inspection_heads});
    \item 
    We find that we can up-weight the value of a few very localized values of the attention head matrix to strengthen factual mechanisms substantially
    (\cref{sec:improving_factual_recall}).
\end{enumerate}

\section{Related  Work on  Interpretability
}
As deep learning approaches show increasingly impressive performance in NLP, their black-box nature has hindered the scientific understanding of these models and their effective future improvements. To this end, interpretability research has been a rising research direction to understand the internal workings of these models.

\paragraph{Interpreting the Representations.}
One major type of work in interpretability has focused on understanding what has been encoded in the representations of deep learning models.
\citep[\textit{inter alia}]{alain2016probe,conneau-etal-2018-cram,hupkes2018visualisation, hewitt-liang-2019-designing,tenney2019WhatDY, jiang-etal-2020-know, elazar2021pararel}, or by geometric methods \cite{doimo2020hierarchical,valeriani2024geometry,park2024geometry,cheng2024emergence}. 
Example features of interest include part of speech \cite{belinkov-etal-2017-neural}, 
verb tense \cite{conneau-etal-2018-cram},
syntax \cite{hewitt-manning-2019-structural}, 
and factual knowledge \cite{petroni-etal-2019-language}.

\paragraph{Interpreting the Mechanisms/Functions.}
Beyond interpreting the representations in the hidden states of the black-box models, another research direction is to interpret the mechanisms or functions that the models have learned, giving rise to the field of mechanistic interpretability
\citep[\textit{inter alia}]{olah2020zoom, elhage2021mathematical, olsson2022incontext,nanda2023grokking}. 
Some example mechanisms decoded in recent work include mathematical operations such as modular addition
\cite{nanda2023grokking} and the greater-than operation \citep{Hanna2023greater-then}; natural language-related operations such as the copy mechanism achieved by induction heads in LLMs \citet{olsson2022incontext} and factual knowledge recall achieved by MLP layers \citep{geva-etal-2021-transformer,meng2022locating,geva2023dissecting}, which we describe below.

\textit{The Single Mechanism of Copy:}
One of the basic actions in LLMs is the copy mechanism, which is found to be operationalized by attending to the copied token in the attention heads and passing it on to the next token prediction \cite{elhage2021mathematical, olsson2022incontext}. This foundational mechanism enables further research to decode more complex mechanisms, such as 
indirect object identification \cite{wang2023IOI}.

\textit{The Single Mechanism of Factual Knowledge Recall:} 
Another major direction is understanding how LLMs mechanistically recall factual information \citep{geva-etal-2021-transformer,meng2022locating,geva2023dissecting}. 
For example, \citet{meng2022locating} develop the \emph{causal tracing} method to show that the factual information is found in the mid-layer MLP units in GPT-2. 
A followup work \cite{geva2023dissecting} shows that MLPs of early layers enrich the subject embeddings with related attributes, and late attention blocks select and write the correct factual information to the sentence's last position.

\textit{Interplay of Multiple Mechanisms:}
In the final stage of our project in December 2023, we noticed a related study by \citet{yu2023characterizing}, which also investigates the role of two different mechanisms in LLMs.
Specifically, they inspect a type of prompt whose subjects are the capital cities and whose attributes are the countries, 
examine the dynamics of the factual recall mechanism and the effect of the in-context counterfactual statement,
and find that the subject and attribute frequency in the pre-training set can affect the ability of factual recall.
Differently, the methods in our work are applied to a broader set of prompts; 
moreover, we also establish novel analyses of the underlying mechanistic details of the competition, and precisely localize the path where the information flows at the level of single attention map activations, based on which we discover new findings that are unique to our study.

\section{Problem Setup}\label{sec:setup}
Following the setup of many existing interpretability studies \citep[\textit{inter alia}]{olah2020zoom, elhage2021mathematical, olsson2022incontext,nanda2023grokking}, we look into the next token prediction behavior of autoregressive LLMs in their inference mode, namely 
\begin{align}
    P(t_k | t_{<k})
    ,
\end{align}
which predicts the $k$-th token $t_k$ given all the previous tokens in the context.

Next, we design the task to incorporate the competition of mechanisms as in \cref{fig:examples}.
Specifically, for each factual statement $\bm{f}:=(t_1^f, \dots, t_k^f)$ consisting of $k$ tokens (e.g., ``iPhone was developed by Apple.''), we compose a corresponding counterfactual statement $\bm{c}:=(t_1^c, \dots, t_{k'}^c)$ (e.g., ``iPhone was developed by Google.''). Then, we compose a prompt connecting the two statements as ``Redefine: $\bm{c}$. $\bm{f}_{1:k-1}$.'', such as \textit{``Redefine: iPhone was developed by Google. iPhone was developed by \_\_\_''}.

The two mechanisms can be traced by inspecting the rise and fall of the factual token $t_k^f$ and the counterfactual token $t_{k'}^c$. For the simplicity of notation, we take the tokens out of the context of their exact position and denote them as $\tfact$ and $\talt$, respectively, in the rest of the paper.

\section{Method and Background}
\label{sec:methods}
\paragraph{Method 1: Logit Inspection.} To inspect the inner workings of the two mechanisms, we trace the \textit{residual stream} \cite{elhage2021mathematical}, or logits of each component in the LLM. 
Given a text sequence of $k$ tokens, LLMs map it
into the residual stream, namely a matrix $\mathbf{x} \in \mathbb{R}^{d \times k}$, where $d$ is the dimension of the internal states of the model.
We use the term $\mathbf{x}_{i}^{l}$ to specify the residual stream at position $i$ and layer $l$. 

An LLM produces the initial residual stream $\mathbf{x}_{i}^{0}$ by applying an embedding matrix $W_E \in \mathbb{R}^{|V| \times d}$ to each token $t_i$, where $|V|$ is the size of the vocabulary. 
Then, it modifies the residual stream by a sequence of $L$ layers, each consisting of an attention block $\mathbf{a}^l$ and MLP $\mathbf{m}^l$. Finally, after the last layer, it projects the internal state of the residual stream back to the vocabulary space with an unembedding matrix $W_U \in \mathbb{R}^{d \times |V|}$.
Formally, the update of the residual stream at the $l^{th}$ layer is:
\begin{equation}
\mathbf{x}^{l} = \mathbf{x}^{l-1} + \mathbf{a}^{l} + \mathbf{m}^{l}
~,
\label{eq:residual_stream}
\end{equation}
where both the attention and the MLP block take as input the $\mathbf{x}$ after layer normalization $\mathrm{norm}$:
\begin{align}
\mathbf{a}^{l} &= \mathbf{a}^{l}(\mathrm{norm}(\mathbf{x}^{l-1}))
~,\label{eq:attention_out} \\
\mathbf{m}^{l} &= \mathbf{m}^{l}(\mathrm{norm}(\mathbf{x}^{l-1}  + \mathbf{a}^{l}))
~.\label{eq:mlp_out}
\end{align}

To understand which token the residual stream $\mathbf{x}^l$ favors, we follow the common practice in previous work \citep{halawi2023overthinking_the_truth, geva2023dissecting, dar2023embeddingspace, geva-etal-2022-transformer} 
to project it to the vocabulary space using the aforementioned unembedding matrix $W_{U}$ which maps the latent embeddings to actual tokens in the vocabulary, enabling us to obtain the logits of the factual $\tfact$ and counterfactual token $\talt$. 

Known as the \emph{Logit Lens} \citep{logitlens2020lesswrong}, this method is broadly adopted due to its consistent success in yielding interpretable results, demonstrating its effectiveness through broad empirical usage. However, it is important to note that it can occasionally fail to reflect the actual importance of vocabulary items, especially in the early layers of the network \citep{belrose2023tunedlens}.

\paragraph{Method 2: Attention Modification.}
Modifying or ablating the activation of a specific model component is also a strategy used to improve the understanding of the information flow within LLMs, 
including techniques such as causal tracing \cite{meng2022locating} and attention knockout 
\citep{wang2023IOI, geva2023dissecting}.

In our work, we focus on modifying a small number of entries in the attention matrix. Namely, in the attention matrix ${A}^{hl}$ of the $h$-th head of the $l$-th attention layer $\mathbf{a}^{l}$, we focus on a certain entry, e.g., at the $(i, j)$ position, where $j<i$, which is the attention value of the token $\mathbf{x}_i^{l}$ attending to one of its earlier tokens $\mathbf{x}_j^{l}$. Following recent work \cite{yu2023characterizing} 
, the modification is after the softmax layer, so the other attention values of the matrix stay unchanged.
For the target entry ${A}_{ij}^{hl}$, we scale it up by a multiplier of $\alpha$:
\begin{equation}
{A}_{ij}^{hl} \leftarrow \alpha \cdot {A}_{ij}^{hl} , \quad \text{ where }j<i~.
\label{eq:alpha}
\end{equation}

\section{Experimental Setup}\label{sec:experimental}

\paragraph{Data Creation}
To compose the factual and counterfactual statements as introduced in \cref{sec:setup}, we adopt 
\textsc{CounterFact}\footnote{\href{https://rome.baulab.info/data/}{https://rome.baulab.info/data/}} \citep{meng2022locating}, commonly used dataset to interpret models' ability of factual knowledge recall. 
We select 10K data points by considering only examples where the attributes are represented by a single token and where the model completes the sentence in a factually accurate manner.

Each instance of \textsc{CounterFact}  expresses a relation $r$ between a subject $s$ and an attribute $a$:  $(s, r, a)$. For example, in the sentence \textit{``iPhone was developed by Apple''}, $s=$ \textit{``iPhone''}, $r=$ \textit{``was developed by''}, $a=$\textit{``Apple''}. Moreover, each $(s, r)$ instance is provided two values of the attribute $a$, namely a factual token $\tfact$, and a counterfactual token $\talt$, representing a false fact.

Using this source data, we compose each instance of our test set in the format of $(\text{\textit{``Redefine:''}}, s,r,{\talt},s,r,\_)$, such as \textit{``Redefine: iPhone was developed by Google. iPhone was developed by \_\_\_''}. We preprocess the original dataset by keeping only the data points whose attribute is a single token (for the simplicity of our implementation), and where the model correctly predicts the factual token $\tfact$ when completing the sentence $(s,r,\_)$. We randomly select 10K test samples into our test set from 219,180 such samples.
We open-source our dataset at {\href{https://huggingface.co/datasets/francescortu/comp-mech}{https://huggingface.co/datasets/francescortu/comp-mech}}.

\begin{figure*}[ht]
    \centering
    \begin{subfigure}{.47\textwidth}
        \centering
        \includegraphics[width=.82\linewidth]{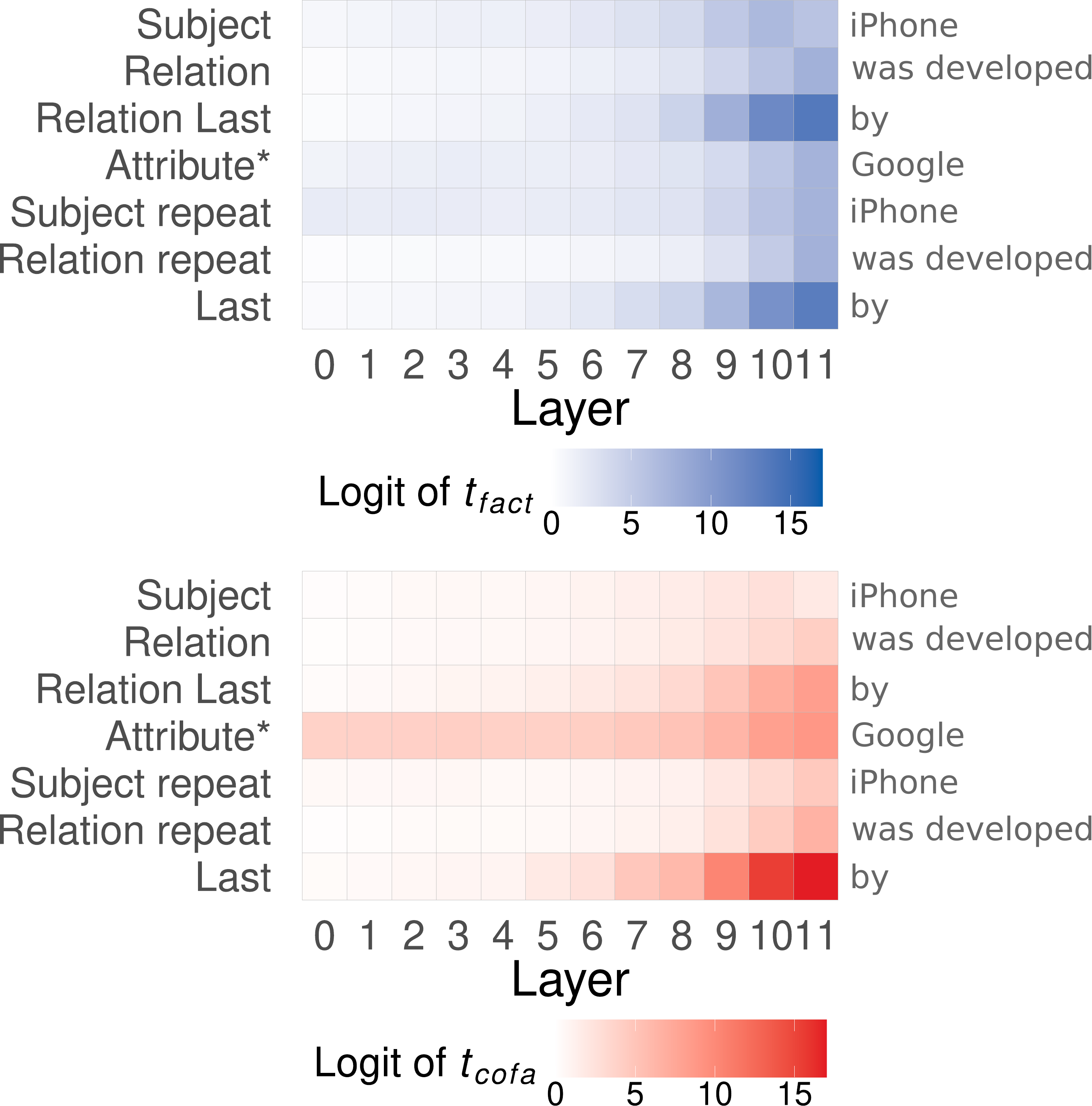}
        \caption{The logit values for the factual token $\tfact$ (blue), and counterfactual token $\talt$ (red).}\label{fig:token_logit}
    \end{subfigure}
    \hfill
    \begin{subfigure}{.47\textwidth}
        \centering
        \includegraphics[width=1\linewidth]{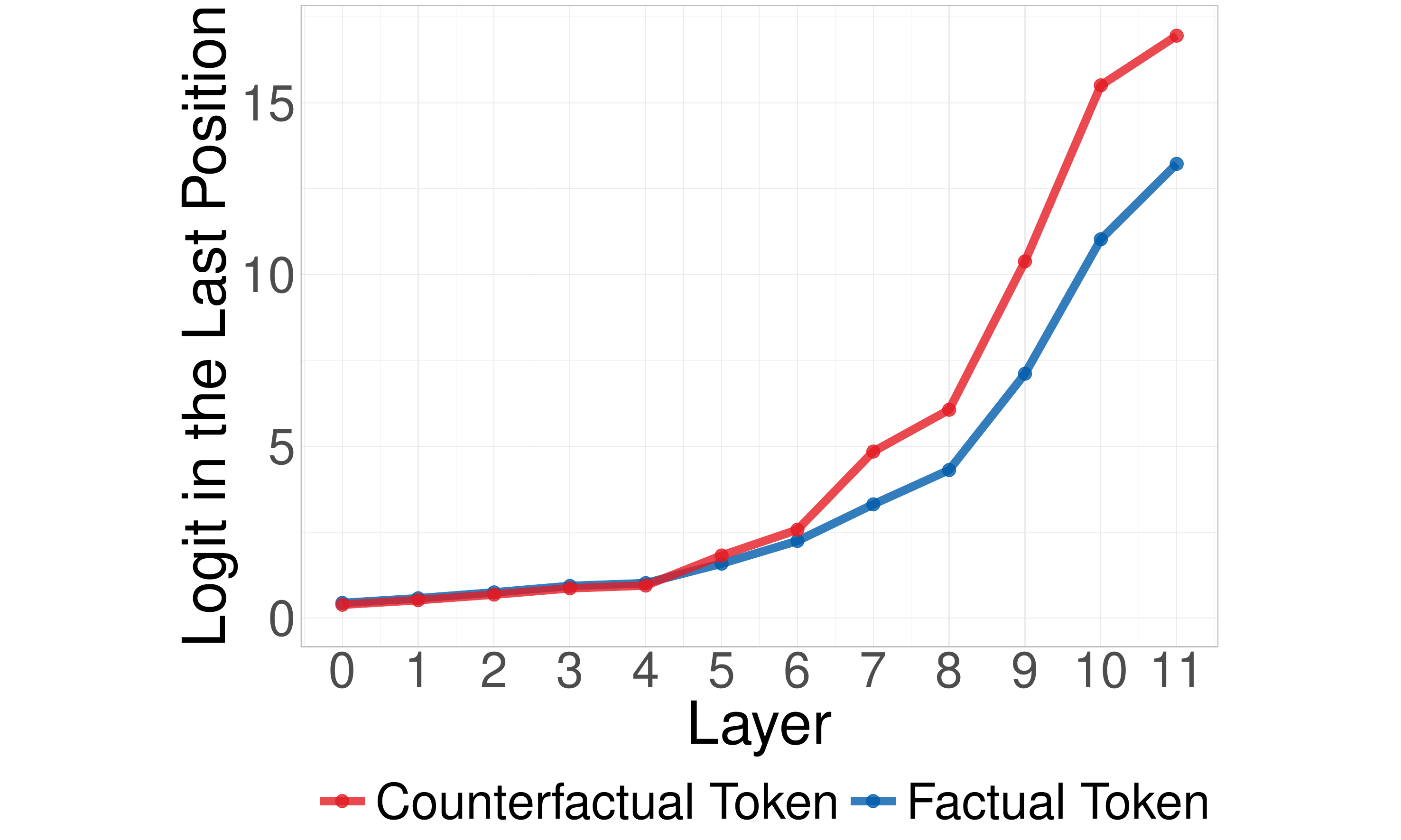}
        \vspace{2em}
        \caption{The average logits of $\tfact$ (blue) and $\talt$ (red) in the last token position.}\label{fig:layer_residual}
    \end{subfigure}
    \caption{
    Logits of the factual token $\tfact$ and counterfactual token $\talt$ across different positions and layers in GPT-2. 
    The logit of $\tfact$ is higher in the subject position in the initial layers and in the last position of the premise and second sentence in the final layers. 
    The logit of $\talt$ is higher in the attribute position in the first layers and in the last position of the second sentence at the end of the network. 
    }
    \label{fig:residual_stream}
\end{figure*}
\paragraph{Models}
For this work, we first choose the GPT-2 small \citep{radford2019language} model as it is the most commonly used one in previous interpretability studies  \citep[e.g.,][]{meng2022locating,wang2023IOI,conmy2023towards_automated_circuit_discovery, Hanna2023greater-then}. Aligning the same model with those studies can communicate the findings of this work better in the context of existing literature. Then, in addition to GPT-2, we check the generalizability of our work by provide supplemental results of Pythia-6.9B \citep{pythia2023} in \cref{appendix:Pythia}, to show the robustness of our findings across the two LLMs of different architectures and scales. In this way, having similar results across the two very diverse models makes the finding stronger than existing studies, most of which are only on GPT-2.

\paragraph{Implementation Details
}
As for experimental details, GPT-2 small has 117M parameters, consisting of 12 layers with 12 self-attention heads each and a residual stream of 768 dimensions.
Pythia-6.9B has 32 layers with 32 self-attention heads each and a model dimension of 4,096, with a 30x increase in the number of parameters.
For all our experiments, we deploy the pre-trained models from the Huggingface Hub \cite{wolf2019transformers}, and inspect the residual streams by the LogitLens tool in the {TransformerLens} library \citep{nanda2022transformerlens}.

\section{Results and Findings}
\label{sec:results}
In this section, we trace the competition of the mechanisms within the LLM via the two methods introduced in \cref{sec:methods}, i.e., inspecting the residual stream and intervening on the attention.
We provide mechanistic analyses on five research questions in the following subsections:
\begin{enumerate}[topsep=0pt]
    \item Macroscopic view: Which layers and token positions contribute to the two mechanisms? (\cref{sec:residual_stream})
\item
Intermediate view: How do we attribute the prediction to attention and MLP blocks? (\cref{sec:attn_mlp_contrbution})
\item
Microscopic view:
How do individual attention heads contribute to the prediction? (\cref{sec:inspection_heads})

\item
Intrinsic intervention:
Can we edit the model activations to modify the strength of a certain mechanism?  (\cref{sec:improving_factual_recall})

\item
Behavioral analysis: What word choice varies the strength of the counterfactual mechanism in the given context? (\cref{sec:similarity_and_competition})

\end{enumerate}

\subsection{Macroscopic Inspection across Layers and Token Positions}
\label{sec:residual_stream}

In the main model that we inspect, GPT-2, we find that it can usually identify the counterfactual mechanism in 96\% of the 10K test examples.
This means that, in the last sequence position, at the output of the network, the counterfactual token, $\talt$, gets most of the times a higher probability than $\tfact$. In the following, we will inspect how the ``winning'' of the counterfactual mechanism happens across the layers of the LLM.

\paragraph{Method.}
We study how $\tfact$ and $\talt$ are encoded in the residual stream using the logit inspection method described in \cref{sec:methods}.
Specifically, for a given token position $i$ and a layer $l$, we project the embedding $\mathbf{x}_i^{l}$, i.e., the residual stream in \cref{eq:residual_stream}, to the vocabulary space by
$\mathbf{\tilde{x}}_i^{l} = W_U \cdot \mathrm{norm}(\mathbf{x}_i^{l}$), where $W_U$ is the unembedding matrix and $\mathrm{norm}$ is the normalization of the last layer of the model. 
By varying $l$, we measure the values of the logits of $\tfact$ and $\talt$ as they evolve in the residual stream after the first attention block.

\begin{figure*}
    \centering
    \begin{subfigure}{.47\textwidth}
    \centering
        \includegraphics[width=0.83\linewidth]{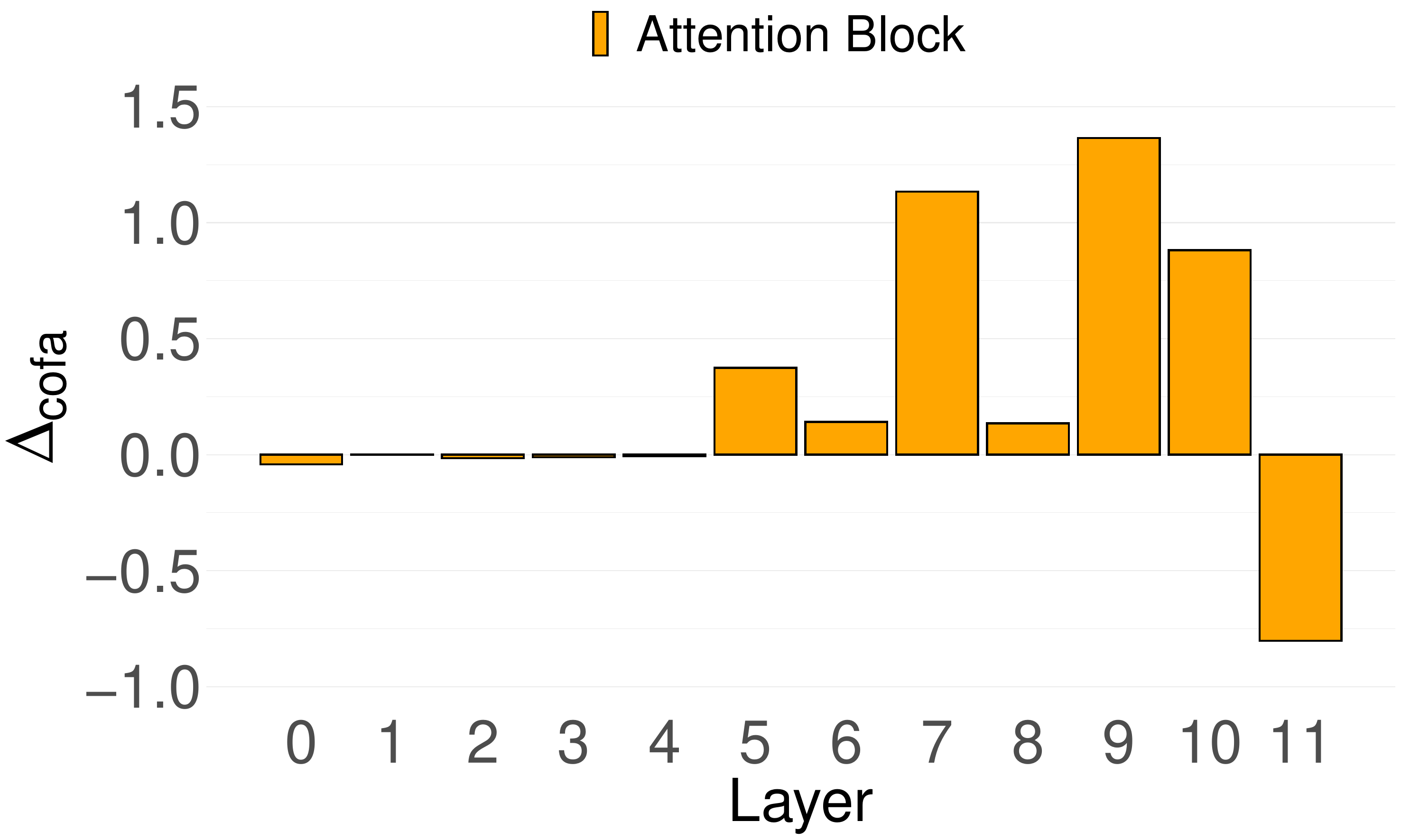}
        \caption{Logit difference $\dalt$ of the last token position after the attention block in each layer of GPT-2.}
\label{fig:attn}

    \end{subfigure}%
    \hfill
    \begin{subfigure}{.47\textwidth}
        \centering
        \includegraphics[width=0.83\linewidth]{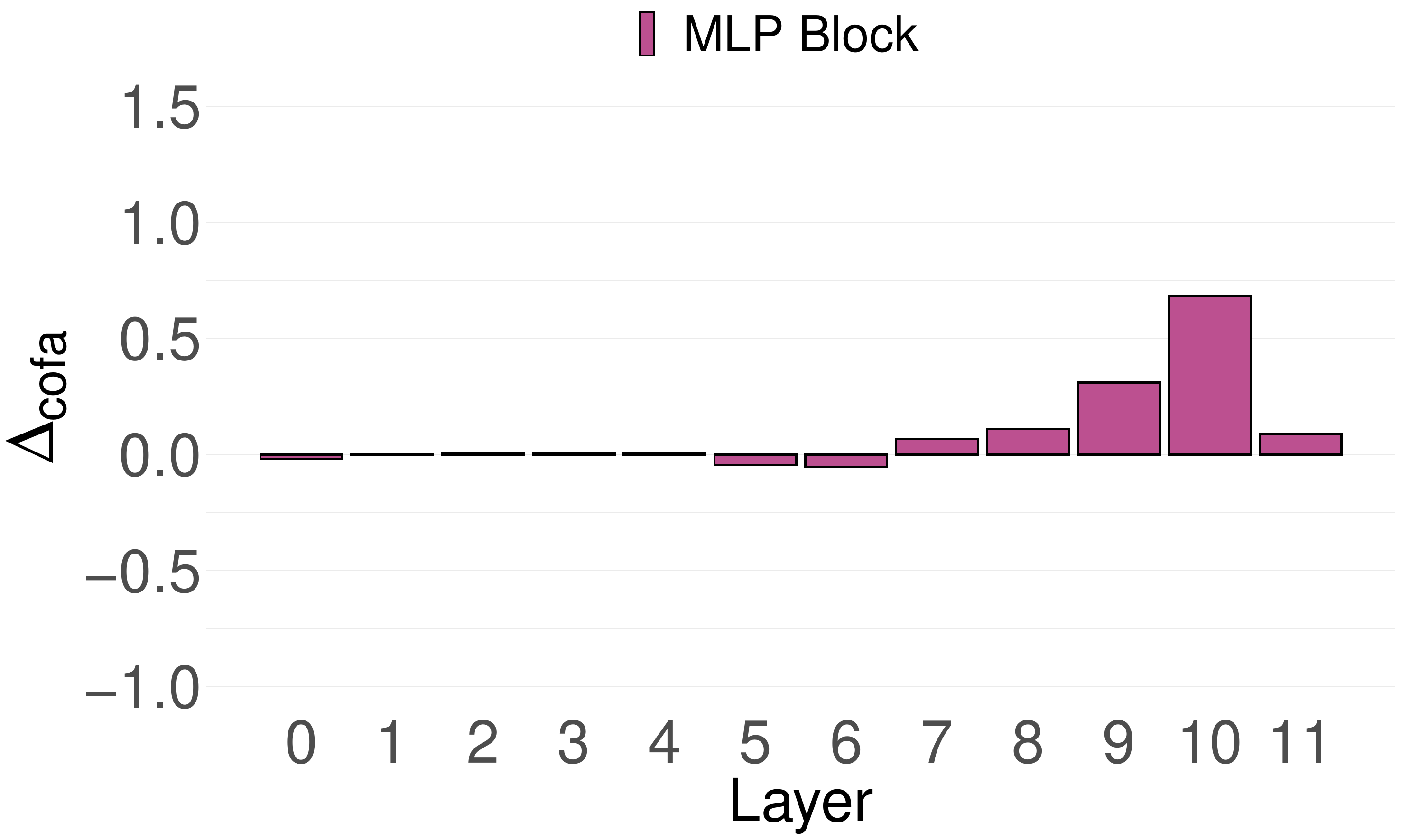}
        \caption{Logit difference $\dalt$ of the last token position after the MLP block in each layer of GPT-2.}
        \label{fig:mlp}
    \end{subfigure}%
    \caption{{Contributions of the attention and MLP blocks to the competition of the mechanisms.} 
    The attention blocks (left) contribute more to the marginal win of the counterfactual mechanism than the MLP blocks (right). 
    }
    \label{fig:attn_mlp_contribution}
\end{figure*}

\paragraph{Results.}
Our results reveal the prevalence of each mechanism by varying the layer $l$ and position $i$.

\textit{Finding 1: Information flows from different tokens for different mechanisms.}
We analyze the role of previous context at various token positions with respect to different depths of the layers. In \cref{fig:token_logit}, the blue heatmap above shows the logits of the factual token $\tfact$, and the red heatmap below shows those of the counterfactual token $\talt$. 

Looking at the blue heatmap, we see that the \textit{subject} position is the main contributor to the logits of $\tfact$ in early layers, which is consistent with a previous finding \cite{geva2023dissecting}. Specifically, we also locate the factual attribute in the subject positions by the first MLP layer, and find they increase on average the value of $\tfact$ from $0.38$ to $0.74$ in the premise and from $0.9$ to $1.93$ in the second sentence.
Then, in the later layers, the strongest contributor is the last tokens before the attribute, as the last token position is used to predict the attribute.
From the red heatmap, we see the evolution of $\talt$'s logits. The observations of later layers are similar across two mechanisms, in that the last token contributes the most.
However, in early layers, the counterfactual mechanism's $\talt$ token is best encoded in the \textit{attribute} position instead of the subject position for the factual mechanism.

Such information flow between different token positions suggests a major role played by the attention mechanism in moving such information to the last position, resonating with observations in \citet{geva2023dissecting}.

\textit{Finding 2: Both the individual mechanisms and competition take place in late, but not early layers.}
We trace the competition of the two mechanisms across the layers by plotting in \cref{fig:layer_residual} the scale of the logits corresponding to the two mechanisms in the last token position.
The first observation is that the strength of each individual mechanism increases monotonically across the layers, from a relatively small logit below 1
in early layers to large values of around 15 in the final layer.

Another observation is that, although both mechanisms increase in strength, stronger signals of the competition (where the counterfactual mechanism prevails the factual one) start after the fifth layer, and this prevalence gradually grows in later layers. The logits of the counterfactual mechanism are,
in most of the examples, the highest in the 50K-dimensional vocabulary of GPT-2, making $\tfact$ dominant in 96\% of the examples.

\subsection{Intermediate Inspection of Attention and MLP Blocks}
\label{sec:attn_mlp_contrbution}
Behind the overall win of the counterfactual mechanism, we want to trace the contributions from the attention and MLP blocks in each layer.

\paragraph{Method.}
For each attention or MLP block, it processes the input embedding and outputs the logits of $\tfact$ and $\talt$ to be added to the residual stream. We can consider the contribution of each block as its added logit values to the residual stream. Intuitively, if the added logit value for $\talt$ is higher than that of $\tfact$, then this block pushes the overall prediction to lean towards the counterfactual mechanism; otherwise, this block suppresses the counterfactual mechanism.

Hence, we inspect the margin of the added logit of $\talt$ over that of $\tfact$ in each block, represented by $\dalt := \text{BlockLogit}(\talt)-\text{BlockLogit}(\tfact)$.
To this end, we apply the logit inspection method to analyze the logit distribution at $W_U \mathbf{a_N}^{l}$ and $W_U \mathbf{m_N}^{l}$, where $N$ denotes the last token position in the sequence.
The logit contribution of the attention block is the sum over that of all the attention heads.
As for the result, a positive value of $\dalt$ for a block means that it supports the counterfactual mechanism in the competition, and a negative value indicates suppression.

\begin{figure*}
\centering
     \begin{subfigure}{.47\textwidth}
        \centering
        \includegraphics[width=0.7\linewidth]{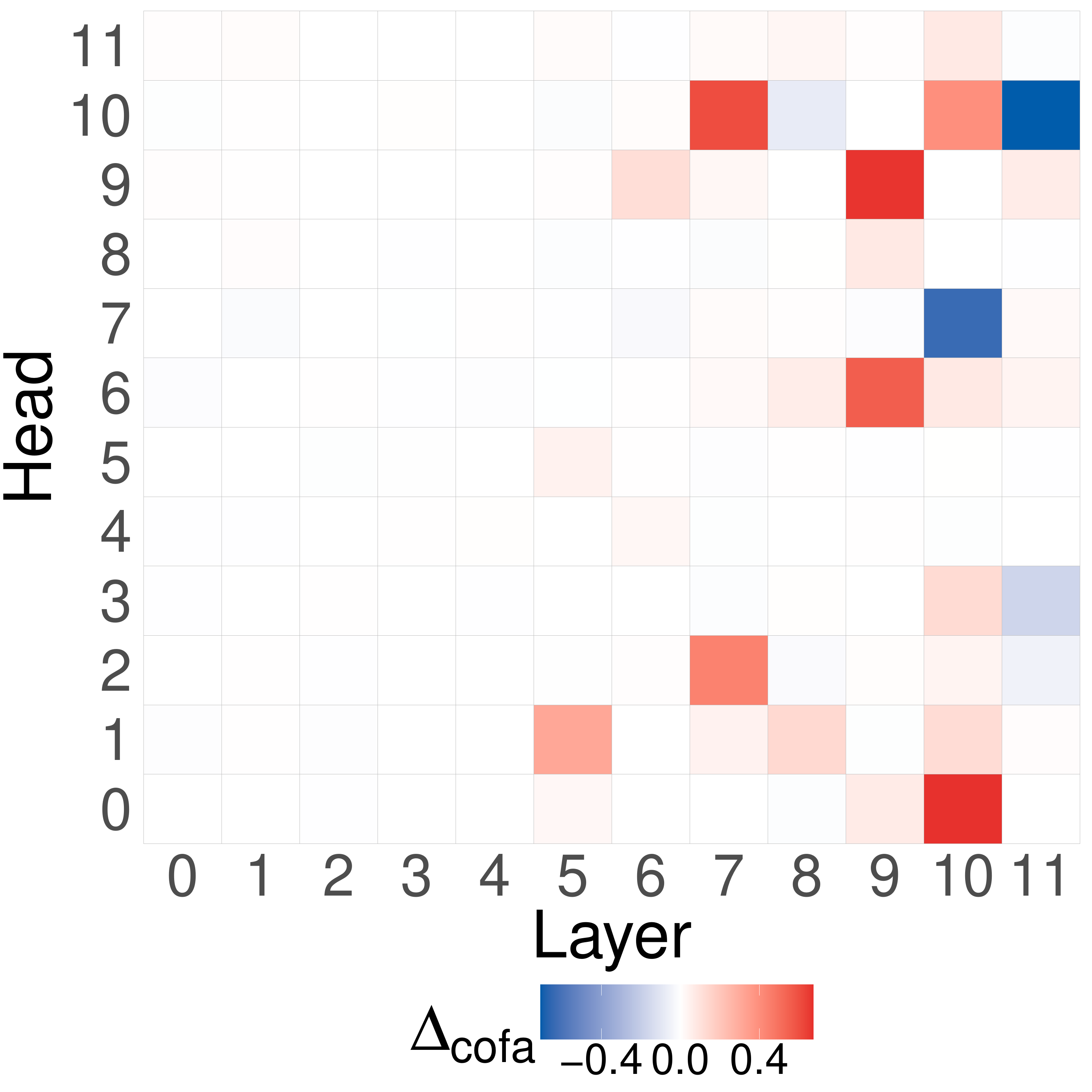}
        \caption{Direct contribution to
    $\dalt$ of all heads in GPT-2. Heads favoring $\tfact$ are colored in blue, and those favoring $\talt$ in red.}
    \label{fig:all_heads}
    \end{subfigure}%
    \hfill
    \begin{subfigure}{.47\textwidth}
    \centering
        \includegraphics[width=0.8\linewidth]{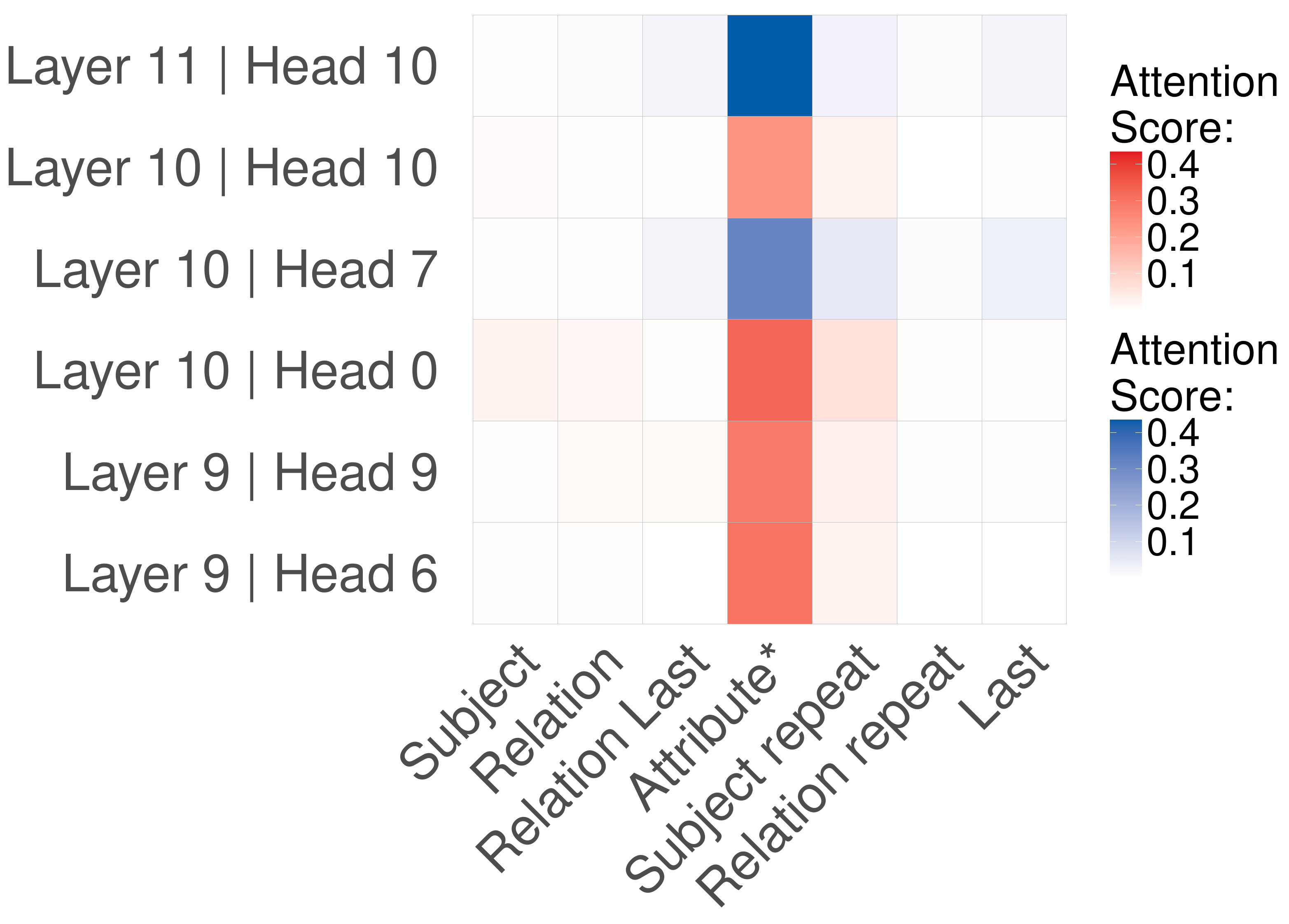}
        \vspace{1.2em}
        \caption{Attention scores of relevant heads for the last position. A large attention score in the attribute position is found in all highly activated heads.}
        \label{fig:main_heads}
    \end{subfigure}%
    \caption{{Attention pattern for relevant attention heads.}}
    \label{fig:attention_heads_analysis}
\end{figure*}
\paragraph{Results.}
We quantify the contribution of each block in each layer by plotting the $\dalt$ values in \cref{fig:attn_mlp_contribution}.

\textit{Finding 1: The attention blocks play a larger role in the competition of mechanisms than the MLP blocks.}
Contrasting the $\dalt$ margin of the added logits of the attention blocks in \cref{fig:attn} and MLP blocks in \cref{fig:mlp}, we see that the size of $\dalt$ is almost always larger in the attention blocks than in MLP blocks. This is consistent with the work of \citet{geva2023dissecting} showing that the attention blocks adds most of the information about $\talt$ to the residual stream.

\textit{Finding 2: Only late but not early layers contribute to the competition of the mechanisms.}
We find that {the early layers have almost no contribution} to the competition of the mechanisms, reflected by the close-to-zero margin $\dalt$ in Layer 0-4 for both types of blocks. However, later layers contribute to substantially to the increase of the margin $\dalt$, by a relatively smaller rate for the MLP blocks, and a larger overall rate for the attention blocks, together with a large variance.

Note that we observe a negative $\dalt$ around -0.8 in the last attention block, somewhat favoring $\tfact$, which might be since the factual information is moved to the last position in the last layers, as already noted by \citealt{geva2023dissecting}.

\subsection{Microscopic Inspection of Individual Attention Heads}
\label{sec:inspection_heads}
Beyond the overall contributions of the attention block, we further study the contribution of each individual attention head in this section.

\paragraph{Method.}
We analyze the effect of each individual attention head with the logit inspection method by projecting the outputs of each attention head to the last sequence position $N$ in the vocabulary space. 
Formally, we consider $\dalt = \text{HeadLogit}(\talt)-\text{HeadLogit}(\tfact)$ with the logits from the projection $W_U \mathbf{a_N}^{h, l}$ of each head $h$. Here $\mathbf{a_N}$ is the output of the attention head $h$ after it has been processed by the output matrix of the Attention Block but before its sum to the residual stream.

\paragraph{Results.}
We plot the contributions of individual attention heads to $\dalt$ in 
\cref{fig:attention_heads_analysis}, and introduce the main findings as follows.

\textit{Finding 1:}
\textit{A few specialized attention heads contribute the most to the competition}. As we can see from the overall contributions of all attention heads across all the layers in \cref{fig:all_heads}, several attention heads (e.g., L9H6, L9H9, L10H0, and L10H10) strongly promote the counterfactual mechanism, i.e., with a positive value of $\dalt$ colored in dark red, and two attention heads (L10H7 and L11H10) strongly support the factual mechanism instead, reflected by the large negative $\dalt$ in dark blue.

For example, the sum of L7H2 and L7H10 equals 75\% of the large positive $\dalt$ contribution of Layer 7. The sum of L9H6 and L9H9 explains 65\% of the $\dalt$ at Layer 9. 

On the other hand, the two attention heads, L10H7 and L11H10, explain almost the 70\% of the total negative contribution to $\dalt$ in the entire network (33\% and 37\% respectively). This also explains the reason behind the negative $\dalt$ in \cref{fig:attn} of the previous section.
Our study is consistent with \citet{mcdougall2023copysupression} showing that these two heads are responsible for suppressing the copy mechanisms in GPT-2 small. In our setting, the joint ablation of these two heads decreases the factual recall of GPT-2 small from 4.13\% to 0.65\%.

\textit{Finding 2: All the highly activated heads attend to the same position -- the attribute token.} 
Focusing on the heads with large absolute values of $\dalt$, we show the attention scores of the last position $N$ to different tokens in \cref{fig:main_heads}. 
Expectedly, the major heads supporting the counterfactual mechanism (those in red) attend to the attribute position because they need to copy this token for the prediction, which echoes the findings in \cref{sec:residual_stream}.

However, it is surprising to see the other heads supporting the factual mechanism (those in blue) also mainly attend to the counterfactual attribute token.
We find that those heads
read from the attribute position
to give a lower value to the logit of $\talt$, which might be an easier operation for it to learn than increasing the logit of the factual token. 
The evidence is that, in these two heads, the logit of $\tfact$ 
is smaller than the mean of the two layers, but the logit of $\talt$ (which is -1.13 for L10H7 and -1.05 for L11H10) are the lowest of all the heads in the network.

We include supplementary analyses showing the consistency of Finding 2 on Pythia in \cref{appendix:Pythia}, and provide
the full attention maps with attention scores between every pair of tokens for these heads in Appendix \cref{app-subsec:full_attn_pattern_gpt2}.

\subsection{Intrinsic Intervention by Attention Modification}
\label{sec:improving_factual_recall}
After \textit{understanding} where the two mechanisms take place, we use the insights to \textit{intervene} on the model internal states. Specifically, we perform model editing to alter the factual mechanism, which concentrates on a few strongly activated attention heads (L10H7 and L11H10 in GPT-2, and mostly L17H28, L20H18, and L21H8 in Pythia, see \cref{app-subsec:attention_heads_pythia}),
and has most of the information flowing from the attribute position (see \cref{fig:attention_heads_analysis}-right and \cref{sec:attn_mlp_contrbution}).
In the following, we show that enlarging the value of a few well-localized attention values can largely improve the factual recall of the model.

\paragraph{Method.}
We utilize the attention modification method in \cref{eq:alpha} to apply a multiplier of
$\alpha$ to the attention weights of the last token to the attribute position in L10H7 and L11H10 for GPT-2, and L17H28, L20H18, and L21H8  in Pythia. 
To choose the $\alpha$ value, we perform a grid search over $[2, 5, 10, 100]$ to maximize the factual recall rate of the model. We find that $\alpha = 5$ is the best value for both GPT-2 for Pythia.

\begin{figure}[t]
    \centering    \includegraphics[width=1\linewidth]{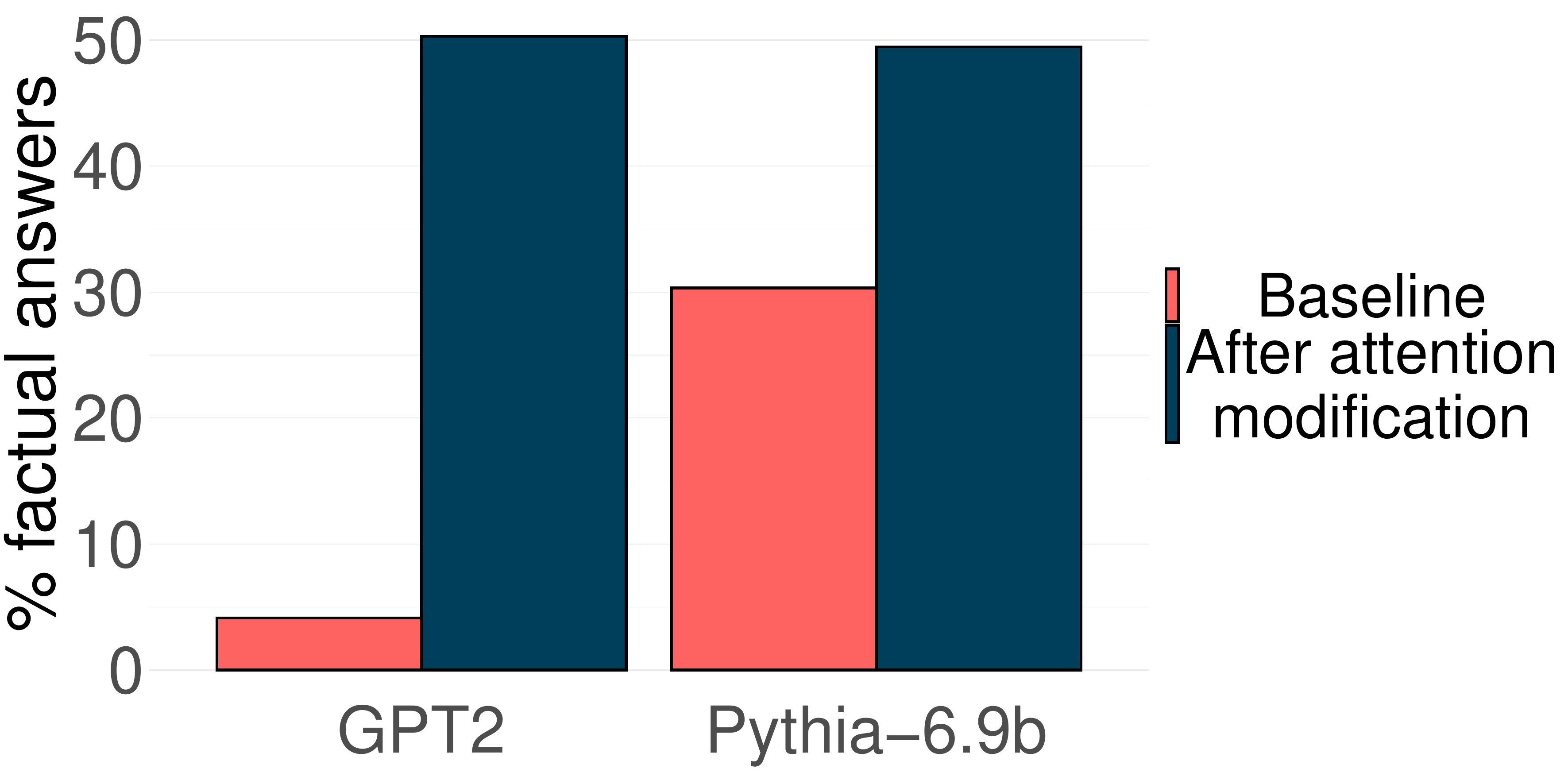}
\caption{{The factual recall mechanism increases substantially across GPT-2 and Pythia after attention modification.} 
}
\label{fig:improved_factual_recall}
\end{figure}

\paragraph{Results.}
We highlight the effect of our model editing method on the strength of the factual recall mechanism in
\cref{fig:improved_factual_recall}.
Originally, GPT-2 has only 4\% of the cases where the factual mechanism prevails the counterfactual one, and Pythia only 30\%. However, after modifying the attention weights of the entries mentioned above, the strength of the factual mechanism increases drastically that it wins over the other mechanism in 50\% of the cases for both models.
This result is remarkable since we modify only two entries in the attention map out of the 33,264 attention values of GPT-2 (117M parameters) and three entries out of the 270,848 attention values of Pythia (6.9B parameters). 
This highlights the importance of the interpretability analysis in \cref{sec:inspection_heads,sec:attn_mlp_contrbution}, which enables us to find the detailed role played by the individual units of each transformer layer.

\subsection{What Word Choices Intensify the Competition?}
\label{sec:similarity_and_competition}
After the intrinsic intervention to edit the internal states of the model, we explore how the similarity between $\tfact$ and $\talt$ in our dataset affects the mechanism described in the previous sections.

\paragraph{Method.}
We divided the dataset into 10 equal bins based on the similarity between the vectors for 
$\tfact$ and $\talt$, with each bin containing $1000$ items. Starting from the lowest, each group represents a $10\%$ segment of the dataset, arranged by increasing similarity scores. %
For our word similarity metric, we calculate the
cosine similarity of the 300-dimensional word embeddings from the pre-trained Word2Vec model \citep{mikolov2013word2vec} implemented in the Gensim Python package \citep{gensim}.
\begin{figure}[t]
    \centering
    \includegraphics[width=0.9\linewidth]{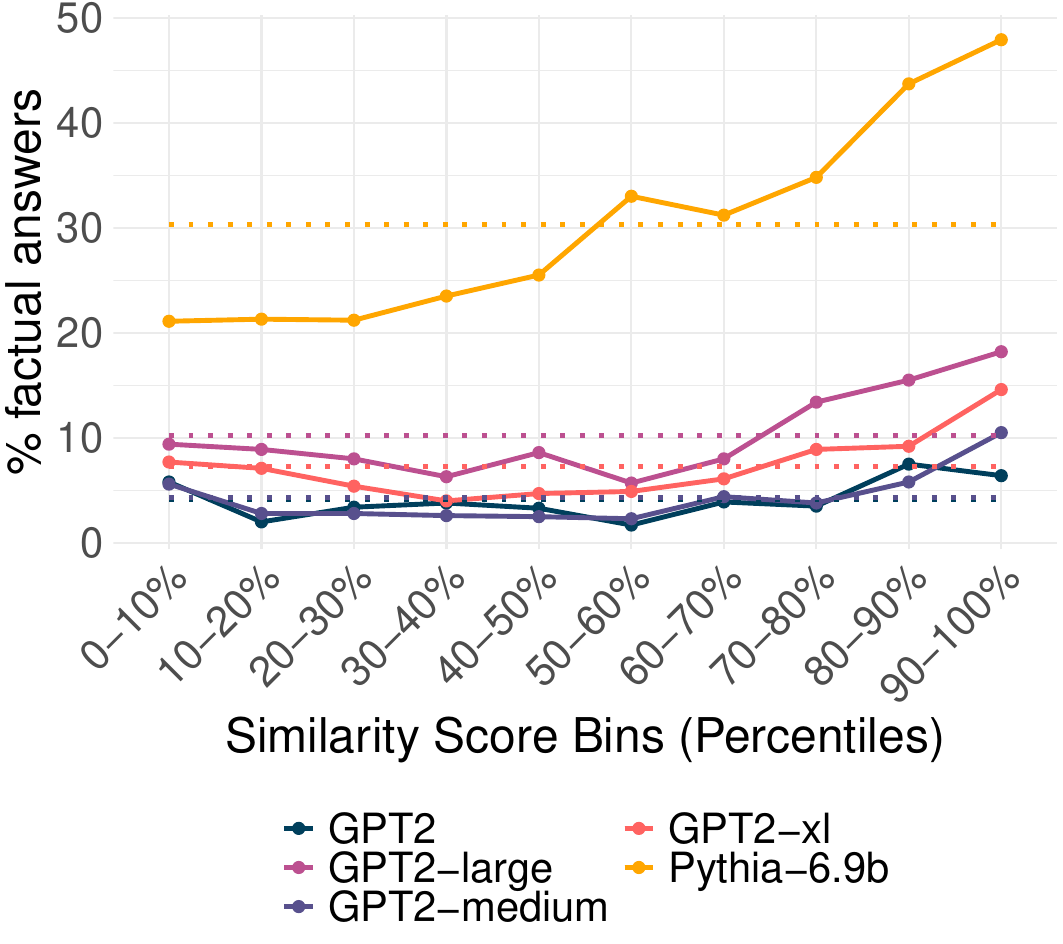}
\caption{{Prediction frequency of factual token by similarity level}. We show the percentage of $\tfact$ predictions within each bin compared to the entire dataset (represented by a dotted line) across various model sizes. 
We can notice that more similar $\tfact$ and $\talt$ are, and the factual mechanism is stronger.}
\label{fig:similarity_semantics}
\end{figure}

\paragraph{Results.}
As a result of the varying similarity of the two tokens, we see a drastic change in the dominance of the factual mechanism in 
\cref{fig:similarity_semantics}.

\textit{Finding 1: Similar tokens confuse the model more easily.}
Consistently across all the models, the more similar the two tokens are, the more likely the model is to be confused and mistakenly let the factual mechanism dominate, to predict $\tfact$ as its output.

\textit{Finding 2:}
\textit{Larger models suffer more from such confusion.} For example, the largest one, Pythia-6.9B, demonstrates a very strong attachment to the factual information, letting the factual mechanism win almost 45\% of the cases when the token similarity reaches 90\%.
Even when the similarity is low, larger models are still more likely to confuse and lean towards the factual mechanism. This finding resonates with the observations from the inverse scaling prize \cite{mckenzie2023inverse} that larger models have a greater capacity to store and retrieve factual information, thus more influenced by the factual mechanism.

\section{Discussion and Future Work}

\paragraph{Situating Our Findings in Related Work.}
Our findings about the late attention blocks are consistent with \citet{geva2023dissecting}, showing that late attention blocks write most of the information to the last layer when adding a counterfactual premise. 
Surprisingly, however, we find that the largest contribution to the factual prediction of the network mostly comes from the suppression of the counterfactual token read from the attribute position rather than the promotion of the factual token from the subject position. 

Consistently with \citet{mcdougall2023copysupression}, we find that few highly specialized heads suppress the counterfactual information. Moreover, we make a unique contribution up-weighting only two or three attention entries of these heads to increase substantially the number of factual responses of the model.

With an approach similar to ours, \citet{yu2023characterizing} find that more heads can promote the factual mechanism, also in early layers, but found it challenging to improve the factual responses by scaling up the weights of the attention maps. 
This discrepancy can be due to the broader set of topics we include in our prompts, which allowed us to select fewer, more specialized heads, to the different ways the prompts are framed, or also to our more focused modification of the attention maps.

\paragraph{Future Work}
For future research directions, we aim to analyze more in depth how our findings depend on the prompt structure and whether the promotion of factual responses by suppressing the counterfactuals generalizes to larger models and a more comprehensive variety of datasets.

\section{Conclusion}
In this work, we have proposed the formulation of the \textit{competition of mechanisms} as a powerful interpretation when LLMs need to handle multiple mechanisms, only one of which leads to the correct answer. We deployed two mechanistic interpretability tools, logit inspection and attention modification, and identified critical positions and model components involved in competing for the mechanisms. Finally, we discovered a few localized positions in the attention map, which largely control the strength of the factual mechanism. Our study sheds light on future work on interpretability research for LLMs.

\section*{Limitations}
\label{sec:limitations}
\textit{Limited models:}
Our study aligns with most existing work in mechanistic interpretability to use GPT-2 small. However, we understand that this is a small model with far fewer parameters than current state-of-the-art LLMs. Future work is welcome to extend to larger-sized models, which might generalize our conclusion to a certain extent, and also reveal interesting behavior once the models get beyond a specific size, maybe also seeing a U-shaped curve \cite{wei2023ushapescaling} for the dominance of the counterfactual mechanism.

\textit{Interpretability method:}
Furthermore, our experiments and insights are heavily grounded in the interpretability within the embedding space of the model's inner components. This approach is reliable and extensively employed in mechanistic interpretability research \citep{dar2023embeddingspace, geva-etal-2022-transformer, halawi2023overthinking_the_truth}.
The logit inspection method, although commonly employed in previous work,
can occasionally fail to reflect the actual importance of some vocabulary items, especially in the early layers of the network \citep{belrose2023tunedlens}.

\textit{Simplicity of the prompts:}
Our prompts have a relatively simple structure for the controllability of the counterfactual information tracing, as it is very challenging to follow the information flow in a more diversified set of prompts. We welcome future work to explore methodological advances to enable analyses over more diverse prompts.

\section*{Ethical Considerations}
The aim of our study is to enhance comprehension of the interplay among mechanisms within language models that may yield unforeseen and undesirable outcomes. Additionally, our research serves as a conceptual demonstration of methods to guide model behavior under such conditions. We believe that recognizing and dissecting the mechanisms by which LLMs produce unpredictable responses is crucial for mitigating biases and unwanted results. Moreover, understanding the competitive dynamics under investigation is critical for improving the safety of LLMs. Specifically, inputting a prompt with an inaccurate redefinition may lead the model to inadvertently reveal sensitive factual information.

\ifarxiv
\section*{Acknowledgment}
We thank Alessandro Stolfo and Yifan Hou for their insightful suggestions, including the pointer to MLP layers for knowledge recall, and many relevant studies. We also thank the audience at the BlackBoxNLP Workshop at EMNLP 2023 for discussions and suggestions on various aspects of this project as well as ethical implications.

This material is based in part upon works supported by the German Federal Ministry of Education and Research (BMBF): Tübingen AI Center, FKZ: 01IS18039B; by the Machine Learning Cluster of Excellence, EXC number 2064/1 – Project number 390727645; 
by the John Templeton Foundation (grant \#61156); by a Responsible AI grant by the Haslerstiftung; and an ETH Grant
(ETH-19 21-1).
Alberto Cazzaniga and Diego Doimo are supported by the project “Supporto alla diagnosi di malattie rare tramite l'intelligenza artificiale'' CUP: F53C22001770002.
Alberto Cazzaniga received funding by the European Union – NextGenerationEU within the project PNRR ''PRP@CERIC'' IR0000028 - Mission 4 Component 2 Investment 3.1 Action 3.1.1.
Zhijing Jin is supported by PhD fellowships from the Future of Life Institute and Open Philanthropy.

\section*{Author Contributions}\label{sec:contributions}

The paper originates as the Master's thesis work of Francesco Ortu hosted jointly at the Max Planck Institute of Intelligence Systems, Tuebingen, Germany, and Area Science Park, Trieste, Italy. 
Zhijing Jin closely supervised the development of the technical idea and the design of the experiments. In the meantime, Francesco developed the technical skills in mechanistic interpretability and conducted the experiments with lots of resilience.

Professors Alberto Cazzaniga and Bernhard Schölkopf co-supervised this work, and gave insightful research guidance.
Diego Doimo closely monitored the execution of the experiments and helped substantially with the design of the word choice experiment and the improvement of the factual recall experiments.
Professor Mrinmaya Sachan provided helpful research suggestions throughout the project.
All of Francesco, Zhijing, Diego, and Alberto contributed significantly to the writing of this paper.

\fi

\bibliography{sec/refs_zhijing,sec/refs_causality,sec/refs_cogsci,sec/refs_nlp4sg,sec/refs_semantic_scholar,refs}

\bibliographystyle{acl_natbib}

\clearpage
\appendix

\section{Experiments for Pythia-6.9b}
\label{appendix:Pythia}
This section extends the experimental analysis conducted on GPT-2 to Pythia-6.9b. The goal is to replicate the prior methodology and compare the outcomes across the two different models, thus contributing to a broader understanding of model behaviors under similar conditions.

\subsection{Macroscopic Inspection across Layers and Token Positions}
\label{app-subsec:residual_stream_pythia}
\begin{figure}[H]
    \centering
    \begin{subfigure}{.4\textwidth}
        \centering
        \includegraphics[width=\linewidth]{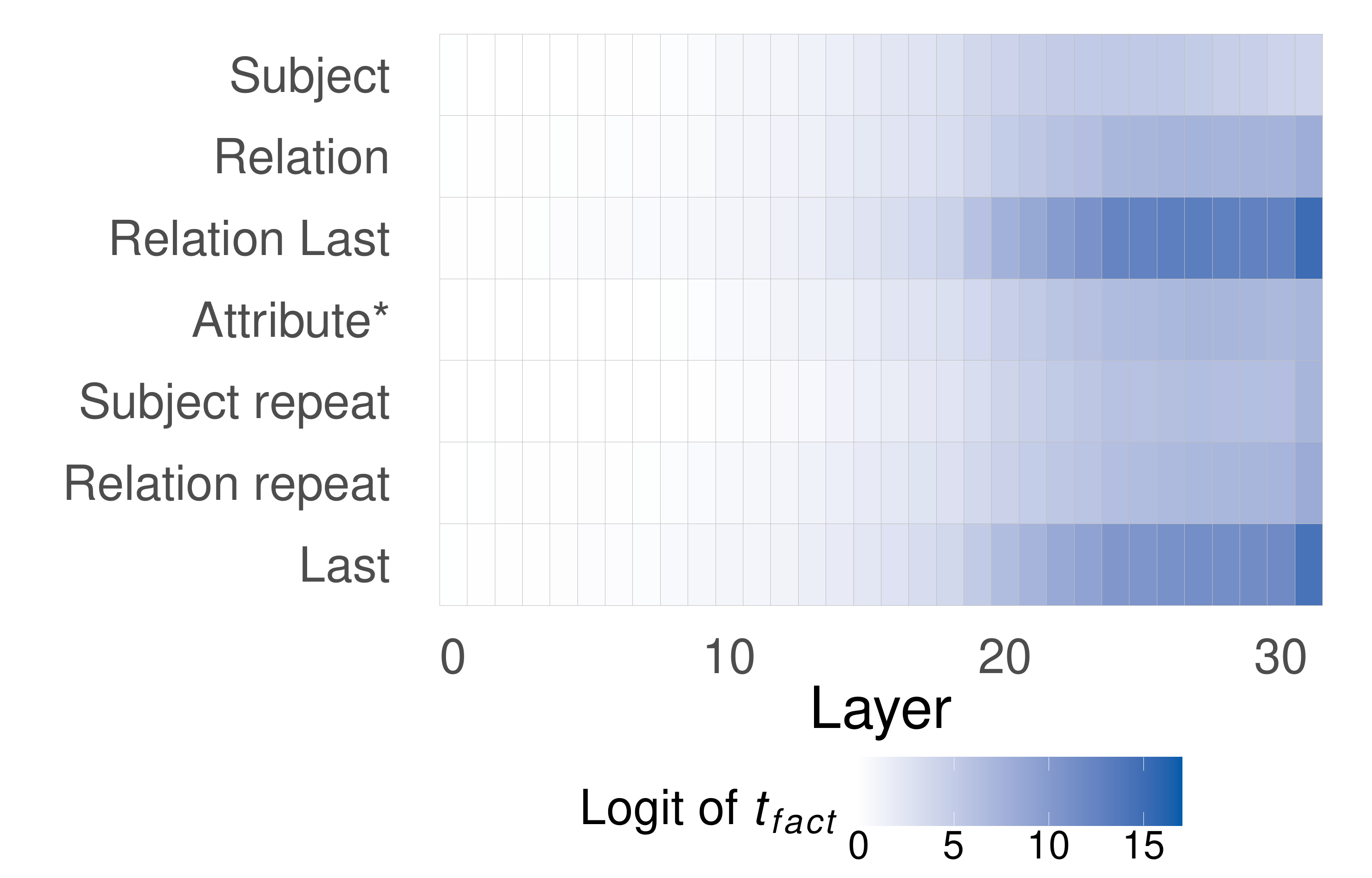}
    \end{subfigure}%
    \hspace{40pt}
    \begin{subfigure}{.4\textwidth}
        \centering
        \includegraphics[width=\linewidth]{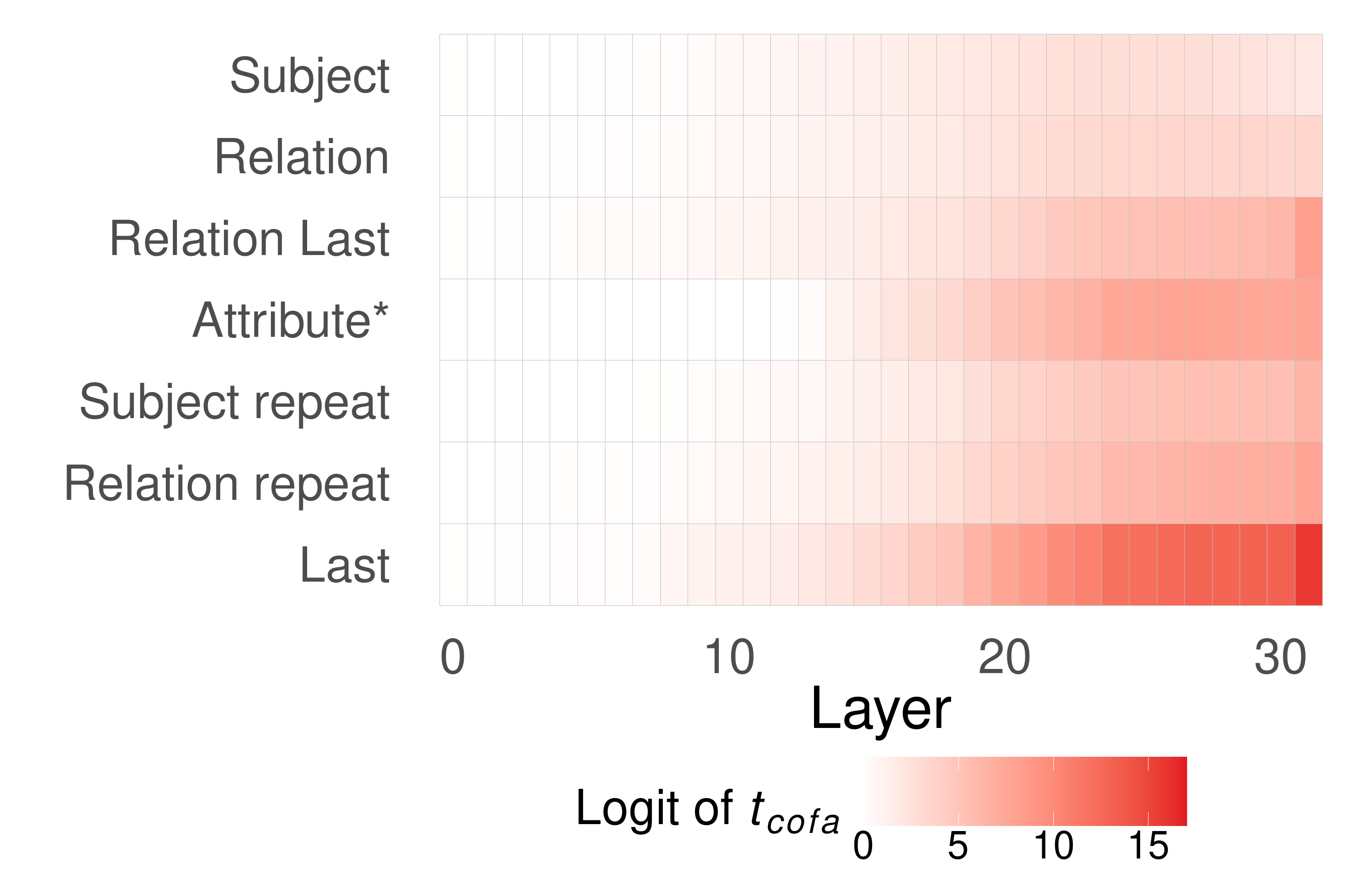}
    \end{subfigure}
    \caption{\textbf{Layer-wise Position Analysis of Relevant Tokens in GPT-2-small}. The figure presents the logit values for two pertinent tokens across various positions and layers. The left panel illustrates the logit values for the factual token $\tfact$, while the right panel illustrates the logit values for the counterfactual token $\talt$. }
    \label{fig:app_residual_stream_pythia}
\end{figure}

\cref{fig:app_residual_stream_pythia} provides a comparative analysis of the logit values for two specific tokens, labeled as factual and counterfactual, across various positions and layers in Pythia-6.9b.

\subsection{Intermediate Inspection of Attention and MLP Blocks}
\label{app-subsec:blocks_pythia}
\begin{figure}[H]
    \centering
    \begin{subfigure}{.46\textwidth}
        \includegraphics[width=\linewidth]{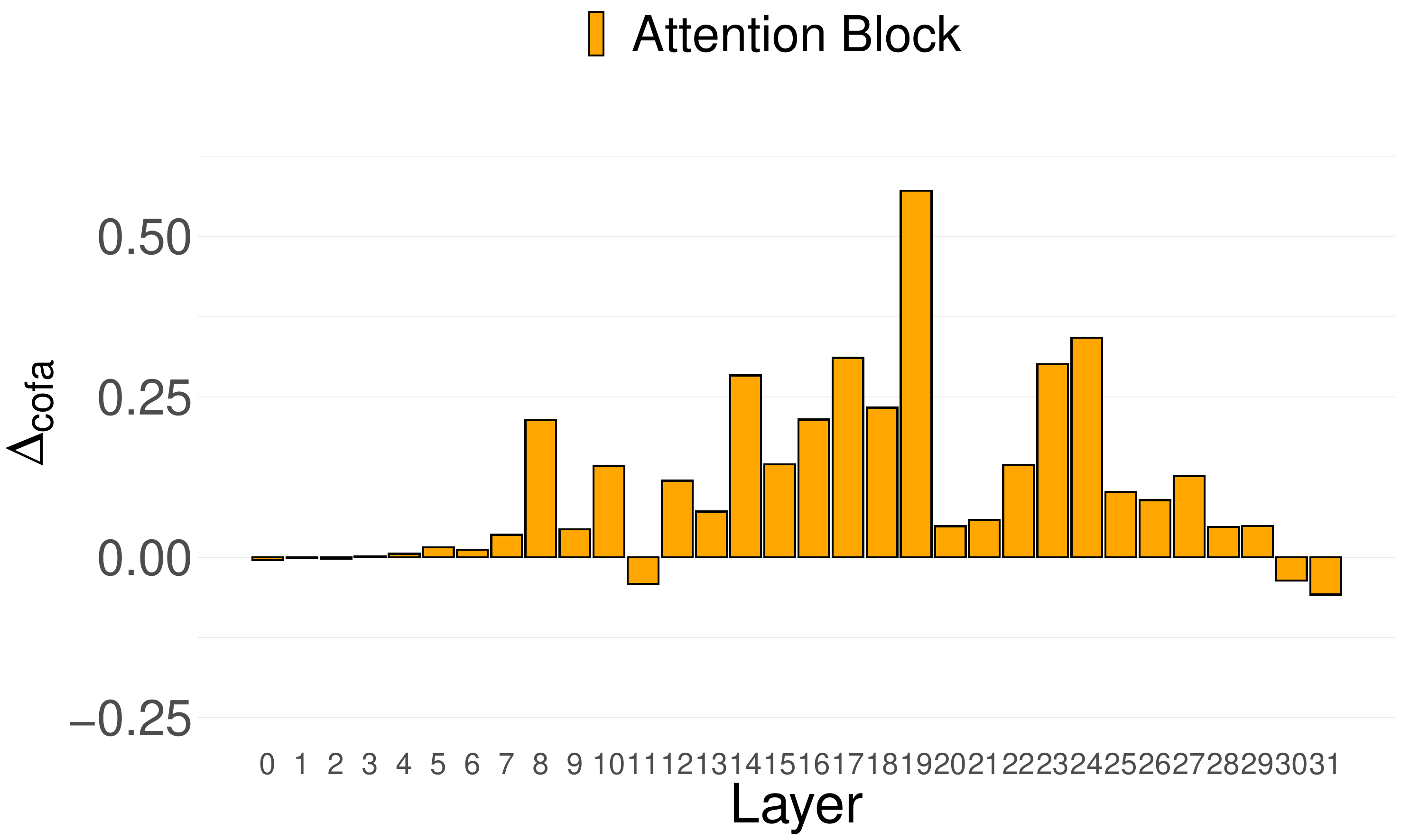}
    \end{subfigure}
    \vspace{10pt}
    \begin{subfigure}{.46\textwidth}
        \centering
        \includegraphics[width=\linewidth]{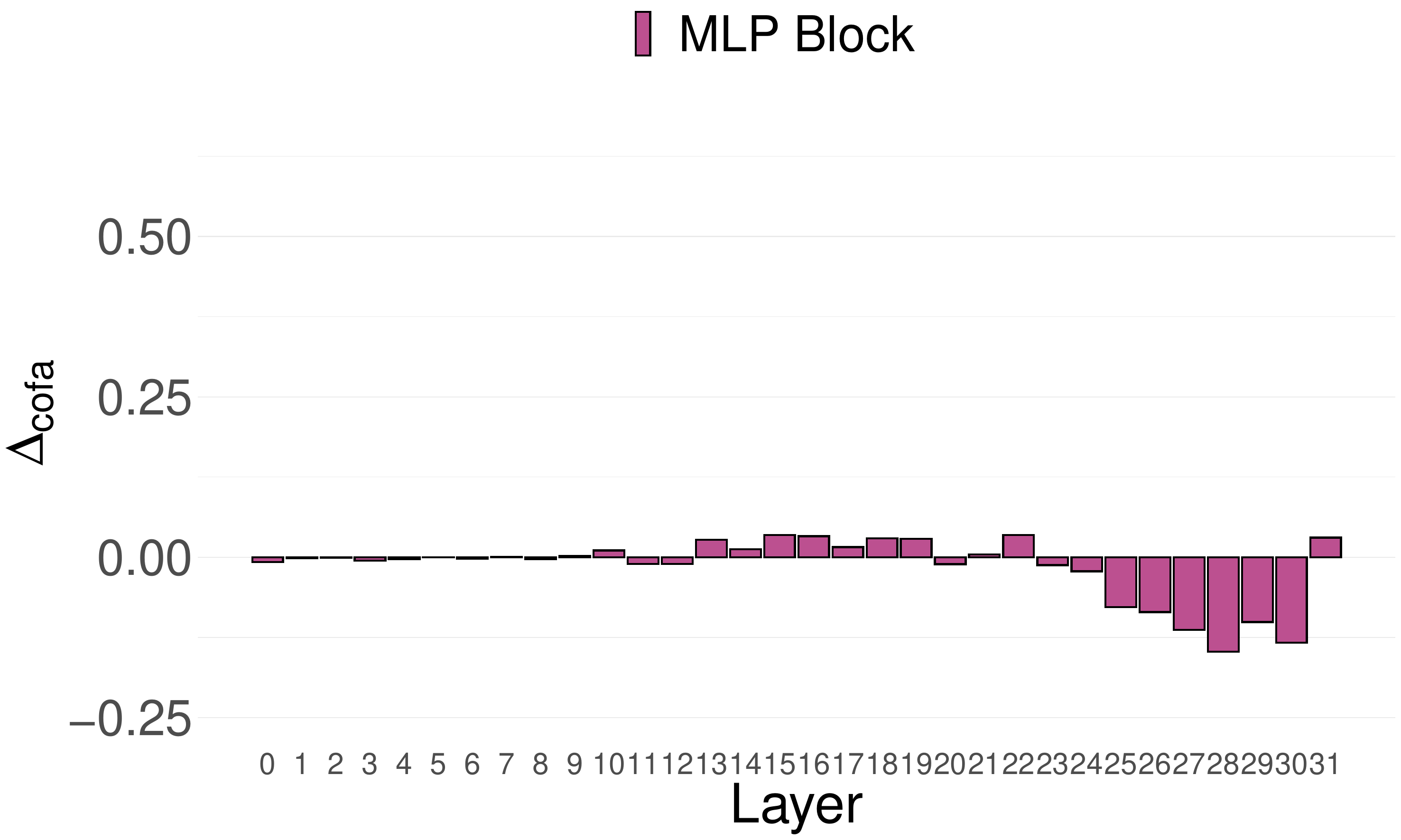}
    \end{subfigure}%
    \caption{\textbf{Attribution of Logit Differences to Attention and MLP Blocks.} delineates the differential impact of Attention and MLP Blocks on logit values at the terminal input position. The attention mechanism is shown to predominantly influence early layer processing in the left panel, while the right panel details the increased contribution of MLP Blocks to the factual token's logits in the concluding layers, illustrating the dynamic interplay between these fundamental neural network elements.}
    \label{fig:app_attn_mlp_contribution_pythia}
\end{figure}

This subsection exposes the contributions of Attention and MLP Blocks to the differences in logit values across layers within Pythia-6.9b. \cref{fig:app_attn_mlp_contribution_pythia} explores how these components influence the computation of logits for two tokens, represented as the difference $\dalt = \text{Logit}(\talt) - \text{Logit}(\tfact)$ at the final position of the input. The analysis specifically highlights the distinct effects of these blocks at different stages of the model's operation.

\subsection{Microscopic Inspection of Individual Attention Heads}
\label{app-subsec:attention_heads_pythia}

\begin{figure*}[h]
\centering

        \includegraphics[width=0.7\linewidth]{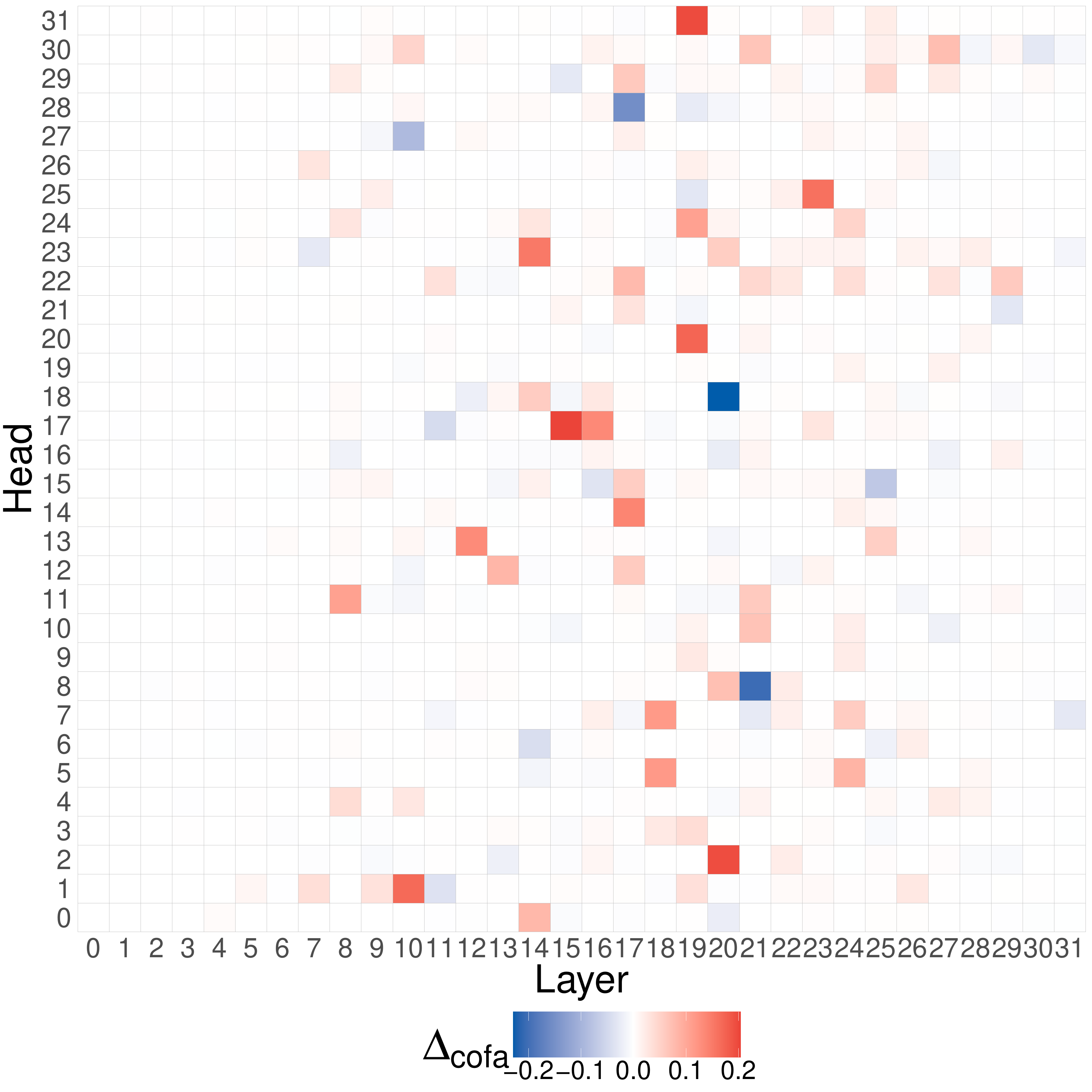}

\caption{\textbf{Direct Contribution of Attention Heads.} The figure displays the direct contribution of all heads in Pythia-6.9b to the logit difference $\Delta_{\text{cofa}}$ with heads favoring $\tfact$ highlighted in blue and those favoring $\talt$ in red.}
\label{fig:app_relevant_attention_heads_pythia}
\end{figure*}

Figure \ref{fig:app_relevant_attention_heads_pythia} quantifies the direct contributions of all attention heads to the difference in logit values, labeled as $\Delta_{\text{cofa}}$. It specifically identifies heads that preferentially enhance the logits for $\tfact$ (shown in blue) versus those favoring $\talt$ (depicted in red), offering insights into how attention mechanisms differentially prioritize token attributes.

\begin{figure}[H]
\centering
        \includegraphics[width=\linewidth]{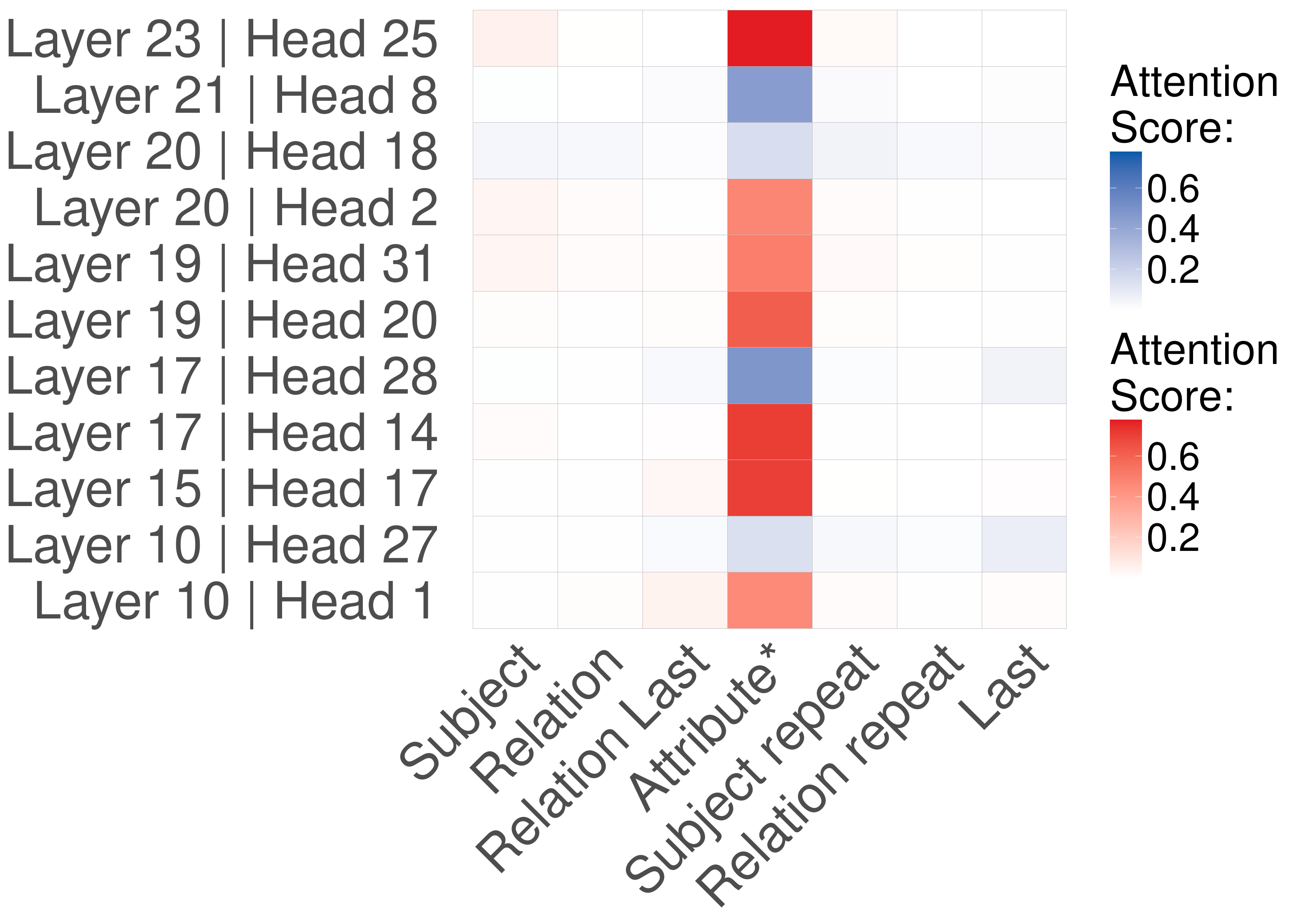}
    \caption{\textbf{Attention Pattern for Relevant Attention Heads.} The panel illustrates the attention patterns of relevant heads for the last position, demonstrating consistent attention to the attribute position by both red and blue heads.  }
    \label{fig:app_attention_heads_analysis_pythia}
\end{figure}

\cref{fig:app_attention_heads_analysis_pythia} presents the attention patterns of the relevant attention heads at the last token position. It shows the consistent pattern of the relevant heads, with a consistent focus on the attribute position.

\section{Other Experiment for GPT-2}
\label{appendix:GPT-2}
\subsection{Ranks Analysis in the Last Position}
\label{app-subsec:rank_gpt2}
\begin{figure}[H]
    \centering
    \begin{subfigure}{.4\textwidth}
        \centering
        \includegraphics[width=\linewidth]{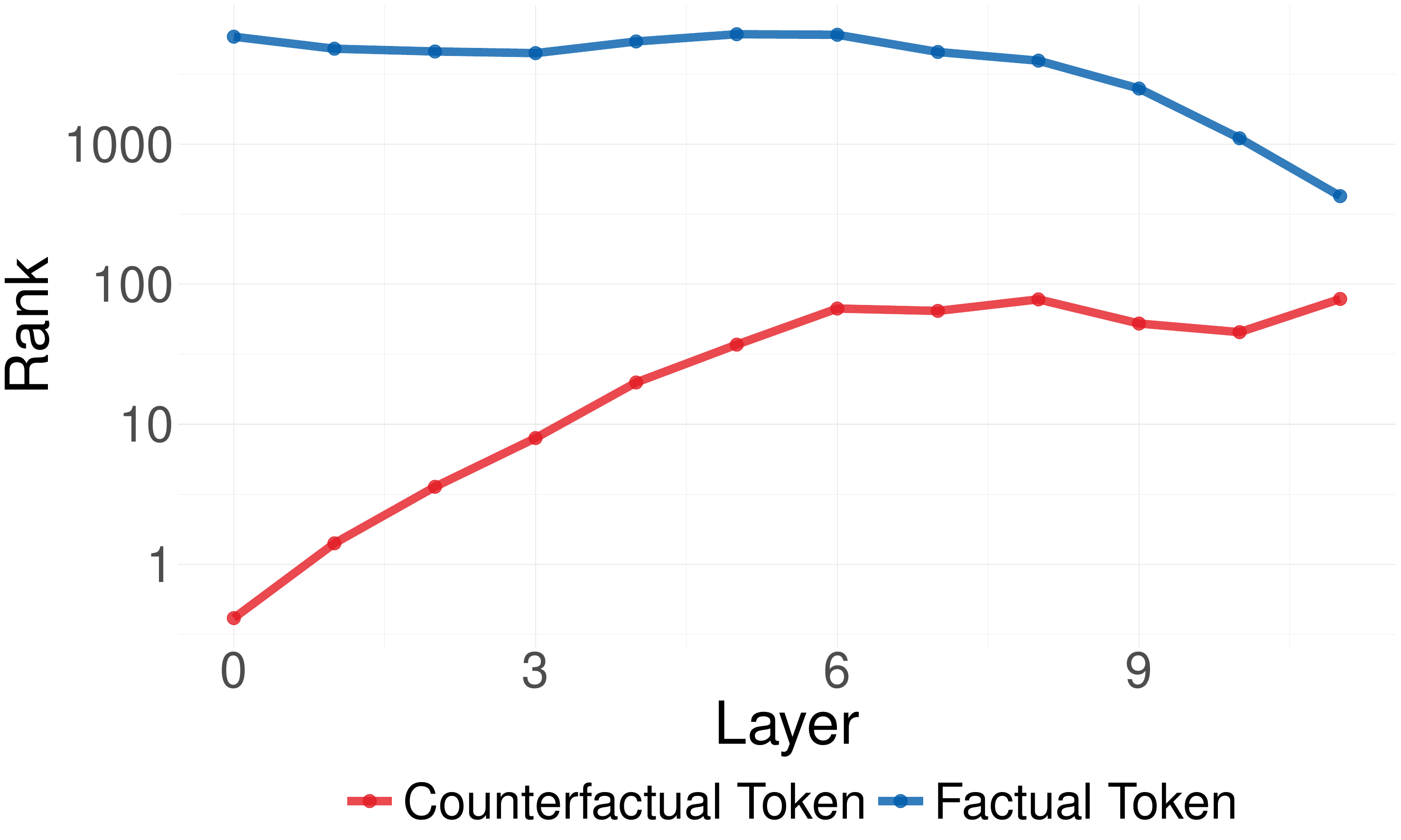}
    \end{subfigure}
    \caption{\textbf{Rank of Target Tokens for Attribute Position Across Layers in GPT-2.} This figure depicts the trend where the logit rank for the factual token $\tfact$ decreases while the rank for the counterfactual token $\talt$ increases at the attribute position. In the concluding layers, this pattern is evident as $\tfact$ typically secures a lower rank, in contrast to $\talt$, which shows an upward trajectory in rank. However, it is important to note that $\talt$'s rank consistently remains lower than that of $\tfact$.}
    \label{fig:app_rank_logit_GPT-2}
\end{figure}

We provide additional information in \cref{fig:app_rank_logit_GPT-2} mapping the logits to ranks of the tokens, and find that 
the rank of $\talt$ in the projected logit distribution remains very low: $\talt$ is among the 20 most likely tokens in the first five layers and between the 20th and the 70th in the last part of the network.
\subsection{Attention Pattern of Relevant Attention Heads}
\label{app-subsec:full_attn_pattern_gpt2}
\begin{figure*}[h]
\centering
     \begin{subfigure}{.8\textwidth}
        \centering
        \includegraphics[width=\linewidth]{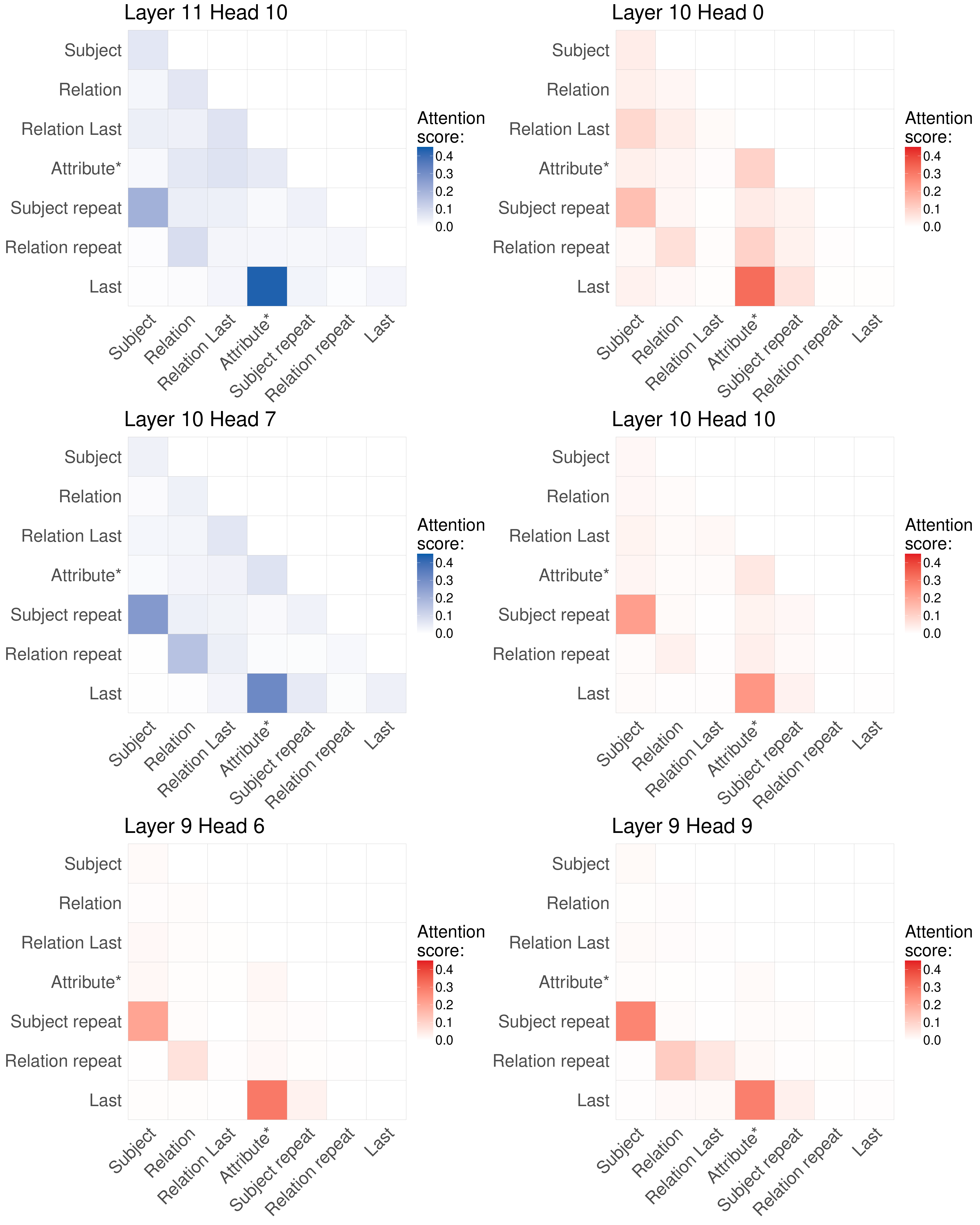}
    \end{subfigure}%
    \caption{\textbf{Attention Pattern of Significant Heads.} This figure illustrates the comprehensive attention pattern of heads substantially influencing $\Delta_{\talt}$. Notably, a similar pattern emerges for both heads favoring $\talt$ (depicted in red) and those favoring $\tfact$ (illustrated in blue), particularly in the attention edge between the attribute and the final position.}
    \label{app:fig_full_head_pattern_GPT-2}
\end{figure*}

\cref{app:fig_full_head_pattern_GPT-2} shows the full attention pattern for the relevant attention heads, as identified in \cref{sec:results}. It is show as the attention pattern is similar between all the relevant attention heads, independently if the heads is favoring $\tfact$ or $\talt$. 
\end{document}